\pdfoutput=1

\documentclass[11pt]{article}

\usepackage{ACL}

\usepackage{times}
\usepackage{bm}
\usepackage{multirow}
\usepackage{xcolor}
\usepackage{colortbl}
\usepackage{mathrsfs}
\usepackage{pifont}
\usepackage{array}

\usepackage{makecell}

\usepackage{algorithm}
\usepackage{algorithmic}
\usepackage{booktabs}
\usepackage{amssymb}
\usepackage{amsmath}
\usepackage{float}
\usepackage{times}  
\usepackage{helvet}  
\usepackage{courier}  
\usepackage{graphicx} 
\usepackage{natbib}  
\usepackage{caption} 
\usepackage{color}

\usepackage{latexsym}
\usepackage{arydshln}

\usepackage[T1]{fontenc}

\usepackage[utf8]{inputenc}

\usepackage{microtype}

\usepackage{inconsolata}
\usepackage{xcolor}
\definecolor{mylightgray}{gray}{0.9}
\definecolor{mygreen}{RGB}{0,150,0}

\title{Taming ``Zombie'' Agents: A Markov State-Aware Framework for Resilient Multi-Agent Evolution}
\author{Taolin Zhang$^{1}$, Pukun Zhao$^{2}$, Qizhou Chen$^{4}$, Jiuheng Wan $^{1}$, Chen Chen$^{2}$, Xiaofeng He$^{4}$, \\ \textbf{Chengyu Wang}$^{3}$\thanks{\ \ C. Wang and R. Hong are co-corresponding authors.}, \textbf{Richang Hong}$^{1}$\footnotemark[1]\\
$^1$ School of Computer Science and Information Engineering, Hefei University of Technology \\
$^2$ Guangdong University of Finance and Economics \\
$^3$ Alibaba Group
$^4$ East China Normal University\\
 {\tt {tlzhang}@hfut.edu.cn, chengyu.wcy@alibaba-inc.com} \\
}

\begin{document}
\maketitle

\begin{abstract}
Recent advancements in LLM-based multi-agent systems have demonstrated remarkable collaborative capabilities across complex tasks. To improve overall efficiency, existing methods often rely on aggressive graph evolution among agents (e.g., node or edge pruning), which risks prematurely discarding valuable agents due to transient issues such as hallucinations or temporary knowledge gaps.
However, such hard pruning overlooks the potential for ``zombie'' agents to recover and contribute in subsequent discussion rounds.
In this paper, we propose \texttt{AgentRevive}, a Markov state-aware framework for resilient multi-agent evolution. Our approach dynamically manages agent collaboration through soft state transitions, implemented via two key components: (1) \texttt{State-Aware Policy Learning}: Agent states are divided into ``\texttt{Active}'', ``\texttt{Standby}'', and ``\texttt{Terminated}'' states, selectively propagating messages based on agent memory. The policy employs a risk estimator to optimize agent state transitions by assessing hallucination risk, minimizing the influence of unreliable nodes while safeguarding valuable ones. (2) \texttt{State-Aware Edge Optimization}: Subgraph edges are pruned according to states learned from the policy, permanently removing ``\texttt{Terminated}'' nodes and retaining ``\texttt{Standby}'' nodes for subsequent rounds to assess their potential future contributions.
Extensive experiments on general reasoning, domain-specific, and hallucination challenge tasks show that our method consistently outperforms strong baselines and significantly reduces token consumption through state-aware agent scheduling.
\end{abstract}

\begin{figure}[!t]
\centering
\includegraphics[width=7cm,height=8cm]{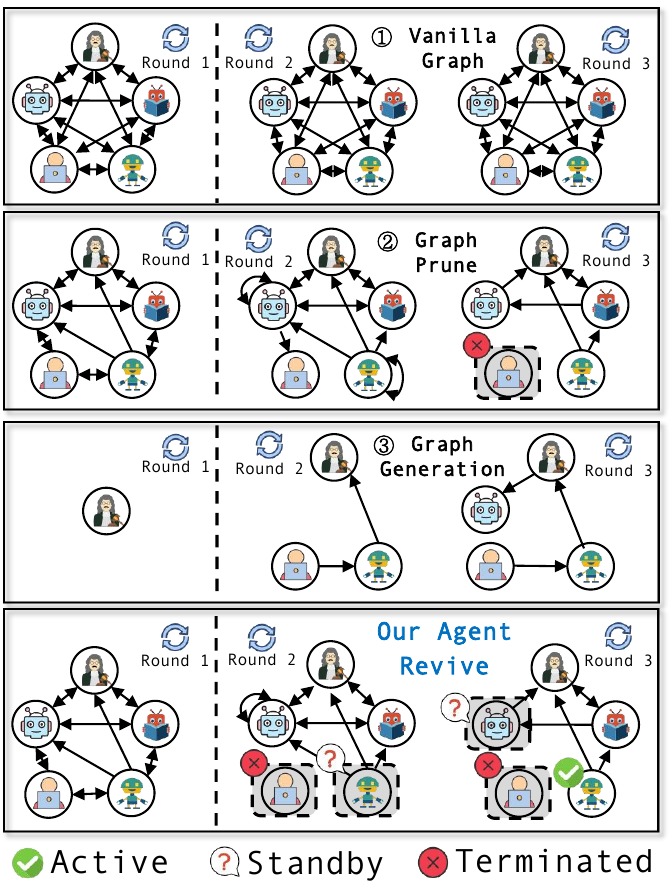}
\caption{Comparison of agent graph topology evolution between our \texttt{AgentRevive} framework and strong training paradigms. (Best viewed in color.)}
\label{motivation_example}
\end{figure}

\section{Introduction}
LLM-powered multi-agent systems (MAS) have emerged as a transformative paradigm for tackling complex tasks, demonstrating superior performance over single-agent methods through collaborative reasoning and planning~\cite{DBLP:conf/acl/00010G00NC25,DBLP:conf/acl/0001LC0H25}.
The efficacy of MAS depends critically on their inter-agent communication topologies, which govern how information is exchanged and assimilated among agents~\cite{DBLP:conf/ijcai/GuoCWCPCW024,DBLP:journals/corr/abs-2502-14321}.
Consequently, recent research has focused on optimizing communication structures to enhance both performance and efficiency~\cite{DBLP:conf/iclr/ZhangYLYWWCY025,DBLP:journals/corr/abs-2506-17784,DBLP:conf/acl/WangW00Z0025}.

Approaches addressing communication redundancy in MAS can be broadly categorized into three paradigms:
\textbf{(1) Vanilla MAS.} These systems rely on manually crafted communication templates, such as chains, trees, or fully connected graphs~\cite{DBLP:conf/nips/Zhang0CPZA24,DBLP:conf/icml/ZhugeWKFKS24,DBLP:conf/naacl/GanZZHLTZS25}. While straightforward to implement, fixed topologies lack adaptability, resulting in inflexible and often inefficient agent interactions that do not dynamically align with task-specific demands.
\textbf{(2) Graph-pruning-based MAS.} This paradigm models agent interactions as graph structures and applies topology-aware learning to prune redundant edges or nodes~\cite{DBLP:conf/acl/WangW00Z0025,DBLP:conf/iclr/ZhangYLYWWCY025,DBLP:journals/corr/abs-2506-02951}. However, this ``hard pruning'' strategy irreversibly removes nodes and edges, potentially discarding useful but temporarily inactive ``zombie'' agents. As a result, the final topology may suffer performance loss, since pruned elements cannot be reactivated even as task contexts change.
\textbf{(3) Graph-generation MAS.} Recent efforts explore autoregressive, dynamic agent graph generation, constructing the collaboration graph from scratch by sequentially generating agent roles and connections~\cite{DBLP:journals/corr/abs-2506-17784,DBLP:journals/corr/abs-2507-18224}.
While this paradigm increases flexibility and avoids initial redundancy, it operates in a purely forward-generative manner without considering the global topological state~\cite{DBLP:conf/iclr/QianXW0ZXDDC00025}.
Consequently, it may fail to reassess or reintegrate previously excluded agents that could become relevant as task conditions evolve, limiting its ability to optimize the communication structure.
As shown in Fig.~\ref{motivation_example}, unlike the three paradigms above, which make permanent pruning decisions, our approach (bottom) allows agent reactivation in later rounds, such as ``Round~3''.

We introduce \texttt{AgentRevive}, a Markov state-aware framework designed for resilient multi-agent evolution.
Our core insight is to treat agent collaboration as a soft, state-aware process rather than relying on hard-pruning decisions.
\texttt{AgentRevive} features two key components:

\begin{itemize}
    \item \textbf{State-Aware Policy Learning}: Learns optimal state transitions for each agent node across communication rounds. We model the agent lifecycle with three states: ``\texttt{Active}'', ``\texttt{Standby}'', and ``\texttt{Terminated}''. State transitions at each round are conditioned on the agent's previous state, its own response, and messages from neighboring agents. To stabilize policy learning under this Markov decision process (MDP), we augment the conventional reward signal, which jointly considers task performance and token efficiency, with a risk estimator. It penalizes strategies that retain agents prone to hallucinated or contradictory responses~\cite{DBLP:journals/corr/abs-2503-13657,DBLP:journals/corr/abs-2505-00212}, encouraging dynamic suspension of unreliable nodes without permanent removal.

    \item \textbf{State-Aware Edge Optimization}: Prunes subgraph edges based on agent states learned from the policy, permanently removing ``\texttt{Terminated}'' nodes and retaining ``\texttt{Standby}'' nodes in subsequent rounds to observe their potential contribution to current tasks.
    Specifically, it constructs a binary node mask based on the survival rates of each agent across multiple inferences, applied to the adjacency matrices of both spatial and temporal edges. This yields a sparsified yet effective communication graph that balances task performance with token efficiency.
\end{itemize}

Experiments across general reasoning, domain-specific, and hallucination benchmarks demonstrate that \texttt{AgentRevive} improves task-averaged performance by \textbf{+2.33\%} compared to strong pruning-based and dynamic autoregressive baselines, while reducing token overhead by \textbf{15\%} through adaptive agent state management.

    

\section{Related Work}

\subsection{Vanilla Agent Collaboration}
Early works demonstrate the effectiveness of single LLM agents in reasoning and planning through structured prompting techniques like chain-of-thought (CoT)~\cite{DBLP:conf/nips/Wei0SBIXCLZ22} and self-consistency (SC)~\cite{DBLP:conf/iclr/0002WSLCNCZ23}.
Subsequent works reveal that MAS can outperform single-agent systems by leveraging specialized capabilities through techniques ranging from majority voting~\cite{DBLP:journals/corr/abs-2403-02419} to sophisticated interaction mechanisms~\cite{DBLP:journals/corr/abs-2308-10848}. 
Recent studies have investigated various predefined communication topologies: (1) \textbf{Non-interactive}: Independent agent operation without interaction, exemplified by LATM~\cite{DBLP:conf/acl/ZhangX0LHD24}, LLM-Blender~\cite{DBLP:conf/acl/Jiang0L23}, and LLM-Debate~\cite{DBLP:conf/icml/Du00TM24}; (2) \textbf{Chain}: Sequential information flow through connected agents, as implemented in ChatDev~\cite{DBLP:conf/acl/QianLLCDL0CSCXL24}, MetaGPT~\cite{DBLP:conf/iclr/HongZCZCWZWYLZR24}, and L2MAC~\cite{DBLP:journals/corr/abs-2310-02003}; (3) \textbf{Star}: Centralized coordination through a commander agent, demonstrated in AutoGen~\cite{DBLP:journals/corr/abs-2308-08155}; (4) \textbf{Tree}: Hierarchical organization with root-level management, such as SoA~\cite{DBLP:journals/corr/abs-2404-02183}. While these predefined templates facilitate effective MAS interaction, they inherently lack flexibility and scalability.

\subsection{MAS Topologies as Graphs}
To improve adaptability, recent approaches have explored learning dynamic communication graphs for MAS from task data.
GPTSwarm~\cite{DBLP:conf/icml/ZhugeWKFKS24} parameterizes agent interactions with DAG topologies optimized via reinforcement learning.
DSPy~\cite{DBLP:conf/iclr/KhattabSMZSVHSJ24} is a programming model that abstracts LLM pipelines as text transformation graphs.
DyLAN~\cite{liu2024} dynamically selects agent teams for task-specific collaboration.
EvoMAC~\cite{DBLP:conf/iclr/HuCDZLYHTC25} employs environmental feedback and textual backpropagation for network updates.
However, these models cannot address redundancy in communication graph structures caused by query-adaptive topology generation.
Graph-pruning-based methods~\cite{DBLP:conf/acl/WangW00Z0025,DBLP:conf/iclr/ZhangYLYWWCY025,DBLP:journals/corr/abs-2506-02951} remove redundant nodes and edges in the temporal and spatial dimensions of the graph based on query-specific characteristics during dynamic topology learning, ultimately forming an adaptive sparse topology for answering the query.
Additionally, autoregressive dynamic graph generation methods~\cite{DBLP:journals/corr/abs-2410-09824,DBLP:journals/corr/abs-2506-17784,DBLP:journals/corr/abs-2507-18224} enable multi-agent pipelines to dynamically generate decision trajectories from scratch, rather than pruning from an initial graph.

\section{Problem Formulation}

In LLM-based multi-agent systems (MAS), agents may temporarily enter a ``zombie'' state, i.e., a failure mode caused by hallucinations or knowledge gaps~\cite{lin2025}. Previous pruning methods~\cite{DBLP:conf/iclr/ZhangYLYWWCY025,DBLP:journals/corr/abs-2506-02951} treat such agents as redundant and remove them permanently. However, if these agents recover in subsequent rounds, they can potentially contribute critically at a later stage.\footnote{Due to space limitations, we refer readers to Appendix~\ref{notations_task} for notations and basic task formulation descriptions.}

To address this, we propose a Markov state-aware collaboration graph framework that dynamically manages agent states across communication rounds.
Specifically, we model MAS as a state-aware collaboration graph $\mathcal{G} = (\mathcal{V}, \mathcal{E}^\mathcal{T}, \mathcal{E}^\mathcal{S}, \mathbf{S})$, where $\mathbf{S}^{(t)} = \{ \mathit{s}_1^{(t)}, \mathit{s}_2^{(t)}, \ldots, \mathit{s}_N^{(t)} \}$ denotes the state of each agent at round $t$, and $\mathit{s}_i^{(t)} \in \{ \texttt{``Active''}, \texttt{``Standby''}, \texttt{``Terminated''} \}$.
The state transition for each agent is governed by a stochastic policy:
\begin{equation}
    \mathit{s}_i^{(t+1)} \sim \pi \left( \cdot \mid \mathit{s}_i^{(t)}, h^{(t)}, m_\mathcal{T}^{(t+1)}, m_\mathcal{S}^{(t+1)} \right)
\end{equation}
where $h^{(t)}$ denotes the interaction history.

We then define the effective subgraph after policy state changes at round $t$ as $\mathcal{G}_{\text{eff}}^{(t)} = (\mathcal{V}_{\text{eff}}^{(t)}, \mathcal{E}_{\text{eff}}^{(t)})$. The effective agent nodes are:
\begin{equation}
    \mathcal{V}_{\text{eff}}^{(t)} = \{ \mathit{v}_i \mid \mathit{s}_i^{(t)} \in \{ \texttt{``Active''}, \texttt{``Standby''} \} \}
\end{equation}
where $\mathcal{E}_{\text{eff}}^{(t)}$ comprises edges between agents in $\mathcal{V}_{\text{eff}}^{(t)}$.
We next reformulate communication redundancy by incorporating agent states~\cite{DBLP:conf/iclr/ZhangYLYWWCY025}.

\noindent \textbf{Definition 1 (State-Aware Redundancy).} Given a state-aware collaboration graph $\mathcal{G} = (\mathcal{V}, \mathcal{E}^\mathcal{T}, \mathcal{E}^\mathcal{S}, \mathbf{S})$, an agent $\mathit{v}_i$ is considered redundant at round $t$ if:
\begin{equation}
    \mathit{s}_i^{(t)} = \texttt{``Terminated''} \quad \text{and} \quad \phi(\mathcal{G}_{\text{eff}}^{(t)}) \geq \phi(\mathcal{G})
\end{equation}
where $\phi(\cdot)$ is a utility function measuring task performance. The state-aware pruning objective is to find a policy $\pi$ that minimizes the effective state-aware graph size while maintaining performance:
\begin{equation}
\label{eq_7}
\min_{\pi} \sum_{t=1}^{T} \left| \mathcal{G}_{\text{eff}}^{(t)} \right|, \quad \text{s.t.} \; \forall t \quad
| \phi( \mathcal{G}_{\text{eff}}^{(t)} ) - \phi( \mathcal{G} ) | \leq \epsilon.
\end{equation}

Table~\ref{design_comparison} summarizes how our Markov state-aware framework offers distinct advantages over conventional graph-based approaches.


\begin{table}[!t]
\footnotesize
\centering
\setlength{\tabcolsep}{3pt}
\begin{tabular}{lccc}
\toprule
\textbf{Method}  & \makecell{\textbf{Task} \\ \textbf{Adaptive}}  &  \makecell{\textbf{Variable} \\ \textbf{Node Size}} &  \makecell{\textbf{Flexible} \\ \textbf{State}}  \\
\midrule
Manual Design & \textcolor{red}{\ding{55}} & \textcolor{red}{\ding{55}} & \textcolor{red}{\ding{55}} \\  
\midrule
AP~\cite{DBLP:conf/iclr/ZhangYLYWWCY025} & \textcolor{mygreen}{\ding{51}} & \textcolor{red}{\ding{55}} & \textcolor{red}{\ding{55}}  \\
G-D~\cite{DBLP:conf/icml/ZhangGUIbin25} & \textcolor{mygreen}{\ding{51}} & \textcolor{red}{\ding{55}} & \textcolor{red}{\ding{55}} \\
AD~\cite{DBLP:conf/acl/WangW00Z0025} & \textcolor{mygreen}{\ding{51}} &  \textcolor{mygreen}{\ding{51}} & \textcolor{red}{\ding{55}} \\
ARG-D~\cite{DBLP:journals/corr/abs-2507-18224} & \textcolor{mygreen}{\ding{51}} & \textcolor{mygreen}{\ding{51}} & \textcolor{red}{\ding{55}} \\ 
\midrule
\texttt{AgentRevive} (Ours)  & \textcolor{mygreen}{\ding{51}} & \textcolor{mygreen}{\ding{51}} & \textcolor{mygreen}{\ding{51}} \\
\bottomrule
\end{tabular}
\caption{Comparison across MAS paradigms. \textcolor{mygreen}{\ding{51}} and \textcolor{red}{\ding{55}} denote full and no support for each capability.}
\label{design_comparison}
\end{table}

\begin{figure*}[!t]
\centering
\includegraphics[width=16cm,height=8.5cm]{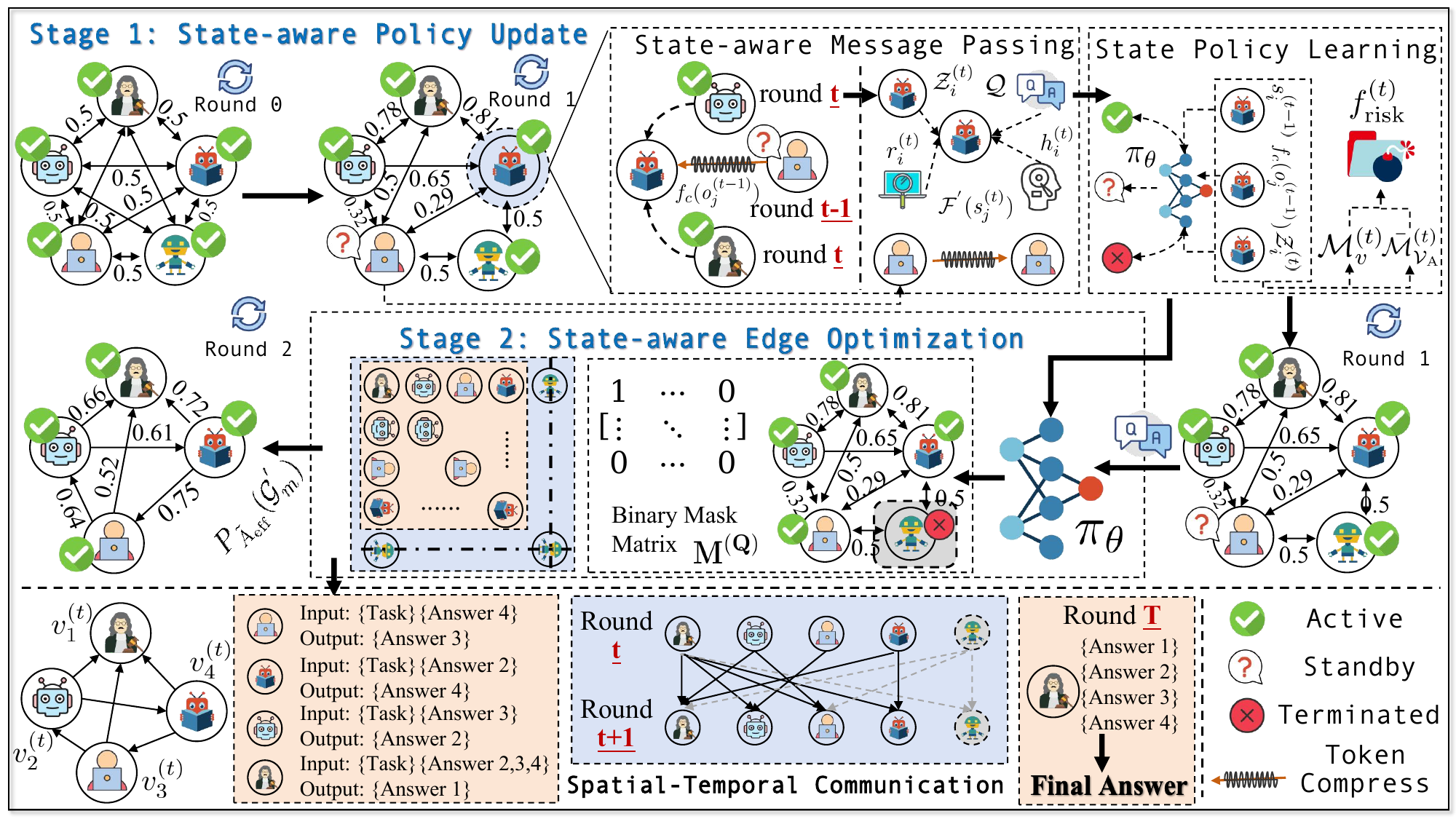}
\caption{Overview of \texttt{AgentRevive}. Our framework mainly consists of two stages for iteratively training: (1) State-Aware Policy Learning is used to aggregate messages around nodes and train agent state policy networks. (2) State-aware Edge Optimization further optimizes the weights of edges around nodes for messages propagation.}
\label{model}
\end{figure*}

\section{Methodology}
\subsection{Notations}

We first convert the state-aware initial collaboration graph $\mathcal{G}$ into a trainable weighted graph $\tilde{\mathcal{G}}$, leveraging pre-defined spatial edges $\mathcal{E}^{\mathcal{S}}$ and temporal edges $\mathcal{E}^{\mathcal{T}}$.
Each edge in the graph is assigned a trainable continuous weight in the range $[0, 1]$.
Let the adjacency matrix set of $\tilde{\mathcal{G}}$ be $\tilde{\mathcal{A}} = \tilde{\mathcal{A}}_{\mathcal{S}} \cup \tilde{\mathcal{A}}_{\mathcal{T}}$, where $\tilde{\mathcal{A}}_{\mathcal{S}} = \bigcup_{t} \tilde{\mathcal{A}}_{\mathcal{S}}^{(t)}$ is the subset containing same-round adjacency matrices, where $\tilde{\mathcal{A}}_{\mathcal{S}}^{(t)} \in [0,1]^{N \times N}$.
$\tilde{\mathcal{A}}_{\mathcal{T}} = \bigcup_{t} \tilde{\mathcal{A}}_{\mathcal{T}}^{(t)}$ is the subset containing temporal-round adjacency matrices, where $\tilde{\mathcal{A}}_{\mathcal{T}}^{(t)} \in [0,1]^{N \times N}$ represents connections between rounds $(t-1)$ and $t$.
The final effective inference graph $\mathcal{G}_{\text{eff}}$ is a DAG, obtained through the learned agent state-aware policy $\pi$.
Fig. \ref{model} is an overview of \texttt{AgentRevive}.

\subsection{State-Aware Policy Learning}
We introduce a paradigm shift from hard pruning to dynamic state management with two key contributions:
(1) State-Aware Message Passing: in contrast to static hard pruning methods ~\cite{DBLP:conf/acl/WangW00Z0025,DBLP:conf/iclr/ZhangYLYWWCY025}, we govern graph message flow using Markov states.
Only agents with the ``\texttt{Active}'' or ``\texttt{Standby}'' state are permitted to propagate messages, forming a dynamic topology that prevents irreversible node removing.
(2) State-aware Policy Decision: the policy determines the optimal state for each node at every iteration. A critical capability is its reactivation of ``zombie'' agents from ``\texttt{Standby}'' to ``\texttt{Active}'' when contextually advantageous ~\cite{lin2025}, thereby ensuring resilience against transient failures.

\subsubsection{State-aware Message Passing}

In our \texttt{AgentRevive} model, the aggregated message $\mathcal{Z}_i^{(t)}$ for agent $v_i^{(t)}$ at round $t$ integrates both spatial and temporal information from the graph $\tilde{\mathcal{G}}$:
\begin{gather}
    \mathcal{Z}_i^{(\mathcal{S},(t))} = \sum_{v_j^{(t)} \in \mathcal{N}_{in}^{\mathcal{S}}(v_i^{(t)})} \tilde{\mathcal{A}}_\mathcal{S}^{(t)}[i,j] \cdot \mathcal{F}(\mathit{s}_j^{(t)}) \\ 
    \mathcal{Z}_i^{(t)} = [\mathcal{Z}_i^{(\mathcal{S},(t))} \parallel \mathcal{Z}_i^{(\mathcal{T},(t))}]
\end{gather}
where $\parallel$ means concatenation. $\mathcal{Z}_i^{(\mathcal{S},(t))}$ is spatial neighboring messages, and the temporal messages $\mathcal{Z}_i^{(\mathcal{T},(t))}$ is also obtained similarly.

After aggregating messages for each agent node, the state-aware response $\mathit{o}_j^{(t)}$ of node $v_j^{(t)}$ at current round $t$ rewrites Eq.~\ref{eq_2} as follows:
\begin{gather}
\mathit{o}_j^{(t)} = 
\begin{cases}
\mathcal{F}'(\mathit{s}_j^{(t)}), & \text{if } \mathit{s}_j^{(t)} = \text{``A''} \\
f_{c}(\mathit{o}_j^{(t-1)}), & \text{if } \mathit{s}_j^{(t)} = \text{``S''}
\end{cases} \\
\mathcal{F}'(\mathit{s}_j^{(t)}) = f_{\text{pr}}\left(r_i^{(t)}, h_i^{(t)}, \mathcal{Q}, \mathcal{Z}_i^{(t)}\right)
\end{gather}
where $\mathcal{F}'(\mathit{s}_j^{(t)})$ is current state response.
$\mathit{o}_j^{(t)}$ is determined by the agent's state: if the state is ``\texttt{Active (A)}'', the current response is used; if the state is ``\texttt{Standby (S)}'', we employ a summarized previous response $f_c(\mathit{o}_j^{(t-1)})$ \footnote{Since $\tilde{\mathcal{A}}$ is an adjacency matrix and $\mathcal{F}^{'}(\mathit{s}_j^{(t)})$ is a string, they are combined via weighted prompts (see Appendix~\ref{implementation_settings}).}.
As historical messages have already been transmitted via different agent nodes in previous rounds, we employ LLM to further compress the number of tokens:
\begin{equation}
\label{eq_9}
f_{c}(\mathit{o}_j^{(t-1)}) = f_{\text{LLM}}(\texttt{``Summarize:''} + \mathit{o}_j^{(t-1)})
\end{equation} 
where $f_{\text{LLM}}$ means using LLM to limit the number of tokens for summary reasoning directly.

\subsubsection{State Policy Learning}

After obtaining the response $\mathit{o}_i^{(t)}$ from state message passing, the model determines the optimal state for agent $v_i^{(t)}$ using its own agent memory:
\begin{equation}
    s_i^{(t)} \sim \pi_\theta(\cdot \mid f_{\text{Enc}}(s_i^{(t-1)}, \mathit{o}_i^{(t)}, h_i^{(t)}, \mathcal{Z}_i^{(t)}))
\end{equation}
where $s_i^{(t)} \in \mathbb{R}^3$.
Given the lightweight nature of our framework, we implement $\pi_\theta$ as a simple MLP for state prediction, and $f_{\text{Enc}}(\cdot)$ utilizes an LSTM~\cite{lstm} to encode the full memory of agent $v_i^{(t)}$, denoted $\mathcal{M}_{v_i}^{(t)}$.

To ensure stable policy training for agent states, we consider policy quality as dependent on both task performance~\cite{DBLP:conf/iclr/ZhangYLYWWCY025} and the level of hallucination present in node responses.
Therefore, the trajectory reward obtained from state transitions of agent nodes is defined as:
\begin{gather}
\label{eq_14}
    R(\tau) = \mu(\tilde{\mathcal{G}}^{(T)}) + \eta_{risk} \cdot \sum_{t=1}^T f_{\text{risk}}^{(t)} \\
    f_{\text{risk}}^{(t)} = - \mathbb{E}_{v \in \mathcal{V}_{\text{A}}^{(t)}} \left[ \mathbb{D}_{\text{KL}} \left(\mathcal{M}_{v}^{(t)} \parallel \bar{\mathcal{M}}_{\mathcal{V}_\text{A}}^{(t)}\right) \right]
\end{gather}
where $\tau$ denotes the trajectory of agent states over rounds $t$; $\mu(\cdot)$ is a utility function measuring the final task score; and $\eta_{risk}$ is a balancing coefficient.
Here, $\mathcal{V}_{\text{A}}^{(t)} = \{ v_i \mid s_i^{(t)} = \text{``A''} \}$ denotes the set of agents in ``\texttt{Active}'' state in round $t$. $f_{\text{risk}}^{(t)}$ quantifies the hallucinatory contradiction of agent node $v_{i}^{(t)}$ in round $t$, using the KL divergence between each agent’s message $\mathcal{M}_{v_i}^{(t)}$ and the average message $\bar{\mathcal{M}}_{\mathcal{V}_\text{A}}^{(t)}$ of all ``\texttt{Active}'' agents.

\subsection{State-aware Edge Optimization}

The strategy trajectory reward focuses on node state transitions. Here, we further consider how these state changes affect the edge weights $\tilde{\mathcal{A}}$, including both spatial and temporal connections. 
Given the learned policy $\pi_\theta^{(\mathcal{Q})}$ for query $\mathcal{Q}$, we re-infer the state matrix $\mathcal{S}^{(t)}$ for each query and calculate the average survival rate $\omega_i$ of each node $v_i$, defined as the proportion of non-``\texttt{Terminated (T)}'' states across $L$ inference passes. This yields a binary node mask vector $\mathbf{m}$ for key node selection:
\begin{gather}
    \omega_i = \frac{1}{L} \sum_{l=1}^L \mathbb{I}(\pi_\theta^{(\mathcal{Q})}(v_i^l) \neq \text{``T''}) \\ 
    \mathbf{m} \in \{0, 1\}^N, \quad m_i = 
        \begin{cases} 
        1 & \text{if } \omega_i \geq \gamma \\ 
        0 & \text{otherwise}
        \end{cases}
\end{gather}
where $\pi_\theta^{(\mathcal{Q})}(v_i^l)$ is the predicted state $s_i^{l}$ at the $l$-th inference for node $v_i$, $\mathbb{I}$ is the indicator function, and $\gamma$ is the survival threshold.

With this mask, we construct the binary mask matrix $\mathbf{M}^{(\mathcal{Q})} = \text{diag}(\mathbf{m}) \in \mathbb{R}^{N \times N}$ and rewrite the state-aware adjacency matrix:
\begin{equation}
    \tilde{\mathcal{A}}_{\text{eff}} = \mathbf{M}^{(\mathcal{Q})} \odot \tilde{\mathcal{A}} \odot \mathbf{M}^{(\mathcal{Q})^\top}
\end{equation}
where $\tilde{\mathcal{A}}_{\text{eff}} \in \mathbb{R}^{N_A \times N_A}$, setting rows and columns for masked nodes to zero vectors. Here, $N_A = \sum_i m_i$ denotes the number of active nodes after removing the ``\texttt{Terminated}'' nodes, and $\odot$ is the element-wise multiplication. The training objective for edge optimization rewrites Eq.~\ref{eq_7}, balancing task performance with graph sparsity:
\begin{equation}
\underset{\tilde{\mathcal{A}}_{\text{eff}}}{\arg\max} \; \underbrace{\mathbb{E}_{\mathcal{G}' \sim \mathbb{G}_{\text{eff}}}[\mu(\mathcal{G}^{'})]}_\text{Performance} - \underbrace{\text{rank}(\tilde{\mathcal{A}}_{\text{eff}})}_\text{Sparsity}
\end{equation}
where $\mathbb{G}_{\text{eff}}$ denotes the feasible domain of graph samples after masking, and $\mu(\mathcal{G}^{'})$ is a task performance evaluation (e.g., APIs), rendering the loss function non-differentiable.
Hence, we employ unbiased policy gradient~\cite{DBLP:journals/ml/Williams92} to approximate this objective, using the weighted average performance of $M$ samples \cite{DBLP:conf/iclr/ZhangYLYWWCY025}:
\begin{multline}
    \nabla_{\tilde{\mathcal{A}}_{\text{eff}}} \mathbb{E}_{\mathcal{G}'\sim \mathbb{G}_{\text{eff}}} [\mu(\mathcal{G}^{'})]  \\
    \approx \frac{1}{M} \sum_{m=1}^M \mu(\mathcal{G}_m^{'}) \nabla_{\tilde{\mathcal{A}}_{\text{eff}}} \log \left( P_{\tilde{\mathcal{A}}_{\text{eff}}} (\mathcal{G}_m^{'}) \right)
\end{multline}
where $P_{\tilde{\mathcal{A}}_{\text{eff}}} (\mathcal{G}_m^{'})$ is the probability of sampling effective subgraph $\mathcal{G}_m^{'} = (\mathcal{V}_m^{'}, \mathcal{E}_m^\mathcal{T}, \mathcal{E}_m^\mathcal{S})$:
\begin{multline}
    P_{\tilde{\mathcal{A}}_{\text{eff}}}(\mathcal{G}_m^{'}) = \prod_{t=1}^{T} \prod_{\left(v_i, v_j\right) \in \mathcal{E}_m^{(t), \mathcal{S}}} \tilde{\mathcal{A}}_{\mathcal{S}}^{\text{eff}, (t)}[i, j] \times \\
    \prod_{t=2}^{T} \prod_{\left(v_i, v_j\right) \in \mathcal{E}_m^{(t), \mathcal{T}}} \tilde{\mathcal{A}}_{\mathcal{T}}^{\text{eff}, (t)}[i, j]
\end{multline}
The second term in the objective constrains the sparsity of the communication graph. To relax the NP-hard rank function, we replace it with the nuclear norm~\cite{DBLP:conf/iclr/ZhangYLYWWCY025}:
\begin{equation}
    \underset{\tilde{\mathcal{A}}_{\text{eff}} = \{\tilde{\mathcal{A}}_\mathcal{S}^{\text{eff}}, \tilde{\mathcal{A}}_\mathcal{T}^{\text{eff}}\}}{\arg\min} \sum_{t=1}^{T} \| \tilde{\mathcal{A}}_{\mathcal{S}}^{\text{eff}, (t)} \|_* + \sum_{t=2}^{T} \| \tilde{\mathcal{A}}_{\mathcal{T}}^{\text{eff}, (t)} \|_*
\end{equation}
where the nuclear norm enables gradient-based optimization of graph sparsity via $\text{rank}(\tilde{\mathcal{A}}_{\text{eff}})$.

\begin{table*}[!t]
\centering
\footnotesize
\setlength{\tabcolsep}{2pt}
\begin{tabular}{ccccccccccc}
\toprule
 \multicolumn{1}{c}{\multirow{1}{*}{\textbf{Dataset}$ \quad \rightarrow$}} &\multirow{2}{*}{ \makecell{\textbf{Var.} \\ \textbf{NS}}} & \multirow{2}{*}{ \makecell{\textbf{Flex.} \\ \textbf{State}}} & \multirow{2}{*}{\textbf{MMLU}} & \multirow{2}{*}{\textbf{GSM8K}} & \multirow{2}{*}{\textbf{AQuA}} & \multirow{2}{*}{\textbf{TruthfulQA}} & \multirow{2}{*}{\textbf{SVAMP}} & \multirow{2}{*}{\textbf{HumanEval}} & \multirow{2}{*}{\textbf{Avg.}} \\
\multicolumn{1}{c}{\multirow{1}{*}{\textbf{Models}$  \quad \downarrow$}}  & & & & & & & & & \\
\midrule
\addlinespace[0.3pt]
\midrule
\multicolumn{10}{c}{Base model: Llama3-8B-Instruct} \\ 
\midrule
\addlinespace[0.3pt]
\midrule

Vanilla & \textcolor{red}{\ding{55}} & \textcolor{red}{\ding{55}} & 53.59 & 70.23 & 41.67 & 57.59 & 75.00 & 53.33 & 58.57 \\ \hdashline
CoT & \textcolor{red}{\ding{55}} & \textcolor{red}{\ding{55}} & 56.86$_{\textcolor{orange}{(\uparrow 3.27)}}$ & 70.47$_{\textcolor{orange}{(\uparrow 0.24)}}$ & 43.75$_{\textcolor{orange}{(\uparrow 2.08)}}$ & 59.25$_{\textcolor{orange}{(\uparrow 1.66)}}$ & 76.17$_{\textcolor{orange}{(\uparrow 1.17)}}$ & 54.17$_{\textcolor{orange}{(\uparrow 0.84)}}$ & 60.11$_{\textcolor{orange}{(\uparrow 1.54)}}$ \\
SC (CoT) & \textcolor{red}{\ding{55}} & \textcolor{red}{\ding{55}} & 60.45$_{\textcolor{orange}{(\uparrow 6.86)}}$ & 71.59$_{\textcolor{orange}{(\uparrow 1.36)}}$ & 46.21$_{\textcolor{orange}{(\uparrow 4.54)}}$ & 59.07$_{\textcolor{orange}{(\uparrow 1.48)}}$ & 78.03$_{\textcolor{orange}{(\uparrow 3.03)}}$ & 55.46$_{\textcolor{orange}{(\uparrow 2.13)}}$ & 61.80$_{\textcolor{orange}{(\uparrow 3.23)}}$ \\ \hdashline
MAS$_{\text{round}=1}$ & \textcolor{red}{\ding{55}} & \textcolor{red}{\ding{55}} & 56.21$_{\textcolor{orange}{(\uparrow 2.62)}}$ & 69.30$_{\textcolor{mygreen}{(\downarrow 0.93)}}$ & 45.29$_{\textcolor{orange}{(\uparrow 3.62)}}$ & 59.88$_{\textcolor{orange}{(\uparrow 2.29)}}$ & 76.67$_{\textcolor{orange}{(\uparrow 1.67)}}$ & 48.33$_{\textcolor{mygreen}{(\downarrow 5.00)}}$ & 59.28$_{\textcolor{orange}{(\uparrow 0.71)}}$ \\
MAS$_{\text{round}=T}$ & \textcolor{red}{\ding{55}} & \textcolor{red}{\ding{55}} & 60.13$_{\textcolor{orange}{(\uparrow 6.54)}}$ & 71.48$_{\textcolor{orange}{(\uparrow 1.25)}}$ & 45.41$_{\textcolor{orange}{(\uparrow 3.74)}}$ & 60.14$_{\textcolor{orange}{(\uparrow 2.55)}}$ & 77.56$_{\textcolor{orange}{(\uparrow 2.56)}}$ & 49.17$_{\textcolor{mygreen}{(\downarrow 4.16)}}$ & 60.65$_{\textcolor{orange}{(\uparrow 2.08)}}$ \\
G-Designer & \textcolor{red}{\ding{55}} & \textcolor{red}{\ding{55}} & 60.27$_{\textcolor{orange}{(\uparrow 6.68)}}$ & 70.59$_{\textcolor{orange}{(\uparrow 0.36)}}$ & 46.82$_{\textcolor{orange}{(\uparrow 5.15)}}$ & 62.43$_{\textcolor{orange}{(\uparrow 4.84)}}$ & 80.03$_{\textcolor{orange}{(\uparrow 5.03)}}$ & 52.53$_{\textcolor{mygreen}{(\downarrow 0.80)}}$ & 62.28$_{\textcolor{orange}{(\uparrow 3.71)}}$ \\
AgentPrune & \textcolor{red}{\ding{55}} & \textcolor{red}{\ding{55}} & 60.78$_{\textcolor{orange}{(\uparrow 7.19)}}$ & 71.02$_{\textcolor{orange}{(\uparrow 0.79)}}$ & 47.22$_{\textcolor{orange}{(\uparrow 5.55)}}$ & 62.83$_{\textcolor{orange}{(\uparrow 5.24)}}$ & 78.34$_{\textcolor{orange}{(\uparrow 3.34)}}$ & 51.67$_{\textcolor{mygreen}{(\downarrow 1.66)}}$ & 61.98$_{\textcolor{orange}{(\uparrow 3.41)}}$ \\
ARG-Designer & \textcolor{mygreen}{\ding{51}} & \textcolor{red}{\ding{55}} & 61.49$_{\textcolor{orange}{(\uparrow 7.90)}}$ & 72.74$_{\textcolor{orange}{(\uparrow 2.51)}}$ & 46.23$_{\textcolor{orange}{(\uparrow 4.56)}}$ & 61.78$_{\textcolor{orange}{(\uparrow 4.19)}}$ & 79.38$_{\textcolor{orange}{(\uparrow 4.38)}}$ & 53.62$_{\textcolor{orange}{(\uparrow 0.29)}}$ & 62.54$_{\textcolor{orange}{(\uparrow 3.97)}}$ \\
AgentDropout & \textcolor{mygreen}{\ding{51}} & \textcolor{red}{\ding{55}} & \underline{62.75}$_{\textcolor{orange}{(\uparrow 9.16)}}$ & \underline{73.13}$_{\textcolor{orange}{(\uparrow 2.90)}}$ & \underline{47.78}$_{\textcolor{orange}{(\uparrow 6.11)}}$ & \underline{63.62}$_{\textcolor{orange}{(\uparrow 6.03)}}$ & \underline{80.11}$_{\textcolor{orange}{(\uparrow 5.11)}}$ & \underline{55.84}$_{\textcolor{orange}{(\uparrow 2.51)}}$ & \underline{63.87}$_{\textcolor{orange}{(\uparrow 5.30)}}$ \\
\rowcolor{mylightgray} \texttt{AgentRevive} & \textcolor{mygreen}{\ding{51}} & \textcolor{mygreen}{\ding{51}} & \textbf{64.30}$_{\textcolor{orange}{(\uparrow 10.71)}}$ & \textbf{75.81}$_{\textcolor{orange}{(\uparrow 5.58)}}$ & \textbf{50.76}$_{\textcolor{orange}{(\uparrow 9.09)}}$ & \textbf{65.49}$_{\textcolor{orange}{(\uparrow 7.90)}}$ & \textbf{82.68}$_{\textcolor{orange}{(\uparrow 7.68)}}$ & \textbf{58.15}$_{\textcolor{orange}{(\uparrow 4.82)}}$ & \textbf{66.20}$_{\textcolor{orange}{(\uparrow 7.63)}}$ \\

\midrule
\addlinespace[0.3pt]
\midrule

\multicolumn{10}{c}{Base model: Deepseek-V3-671B-Instruct} \\

\midrule
\addlinespace[0.3pt]
\midrule

Vanilla & \textcolor{red}{\ding{55}} & \textcolor{red}{\ding{55}} & 84.97 & 94.68 & 84.58 & 64.70 & 93.67 & 88.43 & 85.17 \\ \hdashline
CoT & \textcolor{red}{\ding{55}} & \textcolor{red}{\ding{55}} & 84.31$_{\textcolor{mygreen}{(\downarrow 0.66)}}$ & 95.15$_{\textcolor{orange}{(\uparrow 0.47)}}$ & 85.42$_{\textcolor{orange}{(\uparrow 0.84)}}$ & 64.99$_{\textcolor{orange}{(\uparrow 0.29)}}$ & 93.94$_{\textcolor{orange}{(\uparrow 0.27)}}$ & 89.26$_{\textcolor{orange}{(\uparrow 0.83)}}$ & 85.51$_{\textcolor{orange}{(\uparrow 0.34)}}$ \\
SC (CoT) & \textcolor{red}{\ding{55}} & \textcolor{red}{\ding{55}} & 88.79$_{\textcolor{orange}{(\uparrow 3.82)}}$ & 95.17$_{\textcolor{orange}{(\uparrow 0.49)}}$ & 87.85$_{\textcolor{orange}{(\uparrow 3.27)}}$ & 65.16$_{\textcolor{orange}{(\uparrow 0.46)}}$ & 94.55$_{\textcolor{orange}{(\uparrow 0.88)}}$ & 90.61$_{\textcolor{orange}{(\uparrow 2.18)}}$ & 87.02$_{\textcolor{orange}{(\uparrow 1.85)}}$ \\ \hdashline
AutoGen & \textcolor{red}{\ding{55}} & \textcolor{red}{\ding{55}} & 88.03$_{\textcolor{orange}{(\uparrow 3.06)}}$ & 94.96$_{\textcolor{orange}{(\uparrow 0.28)}}$ & 86.71$_{\textcolor{orange}{(\uparrow 2.13)}}$ & 66.63$_{\textcolor{orange}{(\uparrow 1.93)}}$ & 93.82$_{\textcolor{orange}{(\uparrow 0.15)}}$ & 89.26$_{\textcolor{orange}{(\uparrow 0.83)}}$ & 86.57$_{\textcolor{orange}{(\uparrow 1.40)}}$ \\
AgentVerse & \textcolor{red}{\ding{55}} & \textcolor{red}{\ding{55}} & 87.65$_{\textcolor{orange}{(\uparrow 2.68)}}$ & 95.68$_{\textcolor{orange}{(\uparrow 1.00)}}$ & 85.90$_{\textcolor{orange}{(\uparrow 1.32)}}$ & 65.89$_{\textcolor{orange}{(\uparrow 1.19)}}$ & 94.21$_{\textcolor{orange}{(\uparrow 0.54)}}$ & 88.94$_{\textcolor{orange}{(\uparrow 0.51)}}$ & 86.38$_{\textcolor{orange}{(\uparrow 1.21)}}$ \\
MAS$_{\text{round}=1}$ & \textcolor{red}{\ding{55}} & \textcolor{red}{\ding{55}} & 89.98$_{\textcolor{orange}{(\uparrow 5.01)}}$ & 95.54$_{\textcolor{orange}{(\uparrow 0.86)}}$ & 86.67$_{\textcolor{orange}{(\uparrow 2.09)}}$ & 64.34$_{\textcolor{mygreen}{(\downarrow 0.36)}}$ & 93.50$_{\textcolor{mygreen}{(\downarrow 0.17)}}$ & 89.17$_{\textcolor{orange}{(\uparrow 0.74)}}$ & 86.53$_{\textcolor{orange}{(\uparrow 1.36)}}$ \\
MAS$_{\text{round}=T}$ & \textcolor{red}{\ding{55}} & \textcolor{red}{\ding{55}} & 89.54$_{\textcolor{orange}{(\uparrow 4.57)}}$ & 95.49$_{\textcolor{orange}{(\uparrow 0.81)}}$ & 87.50$_{\textcolor{orange}{(\uparrow 2.92)}}$ & 66.05$_{\textcolor{orange}{(\uparrow 1.35)}}$ & 94.33$_{\textcolor{orange}{(\uparrow 0.66)}}$ & 89.26$_{\textcolor{orange}{(\uparrow 0.83)}}$ & 87.03$_{\textcolor{orange}{(\uparrow 1.86)}}$ \\
G-Designer & \textcolor{red}{\ding{55}} & \textcolor{red}{\ding{55}} & 88.74$_{\textcolor{orange}{(\uparrow 3.77)}}$ & 94.93$_{\textcolor{orange}{(\uparrow 0.25)}}$ & 87.61$_{\textcolor{orange}{(\uparrow 3.03)}}$ & 68.70$_{\textcolor{orange}{(\uparrow 4.00)}}$ & 94.75$_{\textcolor{orange}{(\uparrow 1.08)}}$ & 90.20$_{\textcolor{orange}{(\uparrow 1.77)}}$ & 87.49$_{\textcolor{orange}{(\uparrow 2.32)}}$ \\
AgentPrune & \textcolor{red}{\ding{55}} & \textcolor{red}{\ding{55}} & 90.20$_{\textcolor{orange}{(\uparrow 5.23)}}$ & 95.49$_{\textcolor{orange}{(\uparrow 0.81)}}$ & 87.92$_{\textcolor{orange}{(\uparrow 3.34)}}$ & 69.23$_{\textcolor{orange}{(\uparrow 4.53)}}$ & 95.00$_{\textcolor{orange}{(\uparrow 1.33)}}$ & 90.91$_{\textcolor{orange}{(\uparrow 2.48)}}$ & 88.13$_{\textcolor{orange}{(\uparrow 2.96)}}$ \\
ARG-Designer & \textcolor{mygreen}{\ding{51}} & \textcolor{red}{\ding{55}} & 90.04$_{\textcolor{orange}{(\uparrow 5.07)}}$ & \underline{95.71}$_{\textcolor{orange}{(\uparrow 1.03)}}$ & 87.96$_{\textcolor{orange}{(\uparrow 3.38)}}$ & 68.44$_{\textcolor{orange}{(\uparrow 3.74)}}$ & 94.98$_{\textcolor{orange}{(\uparrow 1.31)}}$ & 91.18$_{\textcolor{orange}{(\uparrow 2.75)}}$ & 88.05$_{\textcolor{orange}{(\uparrow 2.88)}}$ \\
AgentDropout & \textcolor{mygreen}{\ding{51}} & \textcolor{red}{\ding{55}} & \underline{90.85}$_{\textcolor{orange}{(\uparrow 5.88)}}$ & 95.63$_{\textcolor{orange}{(\uparrow 0.95)}}$ & \underline{88.33}$_{\textcolor{orange}{(\uparrow 3.75)}}$ & \underline{70.15}$_{\textcolor{orange}{(\uparrow 5.45)}}$ & \underline{95.79}$_{\textcolor{orange}{(\uparrow 2.12)}}$ & \underline{91.74}$_{\textcolor{orange}{(\uparrow 3.31)}}$ & 88.75$_{\textcolor{orange}{(\uparrow 3.58)}}$ \\
\rowcolor{mylightgray} \texttt{AgentRevive} & \textcolor{mygreen}{\ding{51}} & \textcolor{mygreen}{\ding{51}} & \textbf{91.60}$_{\textcolor{orange}{(\uparrow 6.63)}}$ & \textbf{96.48}$_{\textcolor{orange}{(\uparrow 1.80)}}$ & \textbf{88.85}$_{\textcolor{orange}{(\uparrow 4.27)}}$ & \textbf{72.36}$_{\textcolor{orange}{(\uparrow 7.66)}}$ & \textbf{97.07}$_{\textcolor{orange}{(\uparrow 3.40)}}$ & \textbf{93.52}$_{\textcolor{orange}{(\uparrow 5.09)}}$ & \textbf{90.15}$_{\textcolor{orange}{(\uparrow 4.98)}}$ \\
\bottomrule
\end{tabular}
\caption{Results comparison between \texttt{AgentRevive} and baselines. \textbf{Var. NS} and \textbf{Flex. State} denote the Variable Node Size and Flexible State MAS described in Table~\ref{design_comparison}. The Qwen2.5-72B results are shown in Appendix \ref{other_res}.}
\label{main_res}
\end{table*}

\subsection{Training}

Our \texttt{AgentRevive} model is trained iteratively in two stages. 
The differentiation process for the edge loss function $\mathcal{L}_{\text{edge}}$ in \texttt{Stage 2} is described above. For \texttt{Stage 1}, we compute the state loss $\mathcal{L}_{\text{state}}$ using the REINFORCE algorithm~\cite{DBLP:journals/ml/Williams92}:
\vspace{-0.3cm}
\begin{multline}
    \nabla_\theta \mathcal{L}_{state}(\theta) \approx \\
    \frac{1}{M} \sum_{m=1}^M \sum_{t=1}^T \nabla_\theta \log \pi_\theta(s_i^{(t)} |\mathcal{M}_{v_i}^{(t)}) \cdot (R(\tau) - b)
\end{multline}
where $b$ is a baseline value for variance reduction.
The overall training procedure for \texttt{AgentRevive} consists of first optimizing $\mathcal{L}_{\text{state}}$, followed by optimization of $\mathcal{L}_{\text{edge}}$. Refer to Appendix~\ref{algorithm} for detailed algorithmic training and inference processes.

\section{Experiments}
Due to space limitations, we describe datasets, baselines, and implementation in Appendix~\ref{experimental_details}.

\subsection{Main Results}
Table~\ref{main_res} shows the overall performance in various task settings.
Key observations include: \textbf{[1]} Vanilla methods (CoT~\cite{DBLP:conf/nips/Wei0SBIXCLZ22}, SC~\cite{DBLP:conf/iclr/0002WSLCNCZ23}) outperform standard prompting but show limited gains on complex tasks due to single-agent constraints. \textbf{[2]} Fixed-topology MAS methods (e.g., $\text{MAS}_{\text{round}=1}$, $\text{MAS}_{\text{round}=T}$, AutoGen~\cite{wu2024autogen}, AgentVerse~\cite{DBLP:conf/iclr/ChenSZ0YCYLHQQC24}) underperform single-agent prompting on MMLU~\cite{DBLP:conf/iclr/HendrycksBBZMSS21} and HumanEval~\cite{DBLP:journals/corr/abs-2107-03374}, due to inefficient communication overhead~\cite{DBLP:journals/corr/abs-2505-13466,DBLP:conf/acl/0002JRPYWHX0D025} and error propagation in multi-round setups.
\textbf{[3]} Graph-based dynamic MAS methods improve over vanilla MAS via adaptive topologies, though hard pruning risks the permanent loss of potentially useful agents.
\textbf{[4]} \texttt{AgentRevive} achieves SOTA results, with notable gains on TruthfulQA~\cite{DBLP:conf/acl/LinHE22}, by dynamically managing agent states to mitigate hallucinations while preserving recovery potential.\footnote{The detailed analysis is shown in Appendix~\ref{other_res}.}

\begin{table}[!tb]
\centering
\footnotesize
\setlength{\tabcolsep}{0.5pt}
    \begin{tabular}{lcccc}
      \toprule
\multicolumn{1}{c}{\multirow{1}{*}{\textbf{Dataset}$ \quad \rightarrow$}} & \multirow{2}{*}{\textbf{MMLU}} & \multirow{2}{*}{\textbf{GSM8K}} & \multirow{2}{*}{\textbf{TQA}}  & \multirow{2}{*}{\textbf{Avg.}} \\
\multicolumn{1}{c}{\multirow{1}{*}{\textbf{Models}$  \quad \downarrow$}}  & & & & \\
    
\midrule
\addlinespace[0.3pt]
\midrule
\multicolumn{5}{c}{Base model: Llama3-8B-Instruct} \\ 
\midrule
\addlinespace[0.3pt]
\midrule

  \rowcolor{mylightgray} \multicolumn{1}{c}{     $\textbf{AgentRevive}$}   & \textbf{64.30} & \textbf{75.81} & \textbf{65.49} & \textbf{68.53}   \\ \midrule
    \multicolumn{1}{c}{w/o SEO }   & 62.08 &73.59 & 62.37 & 66.01  \\
     \multicolumn{1}{c}{ w/o SPL  } & 59.27$_{\textcolor{red}{(\downarrow 5.03)}}$ & 69.85$_{\textcolor{red}{(\downarrow 5.96)}}$ & 59.28$_{\textcolor{red}{(\downarrow 6.21)}}$ & 62.80$_{\textcolor{red}{(\downarrow 5.73)}}$ \\
     \multicolumn{1}{c}{ w/o SMP } & 61.25 & 71.81 & 60.79 & 64.62  \\ \midrule
\addlinespace[0.3pt]
\midrule
\multicolumn{5}{c}{Base model: Qwen2.5-72B-Instruct} \\ 
\midrule
\addlinespace[0.3pt]
\midrule
  \rowcolor{mylightgray} \multicolumn{1}{c}{     $\textbf{AgentRevive}$}   & \textbf{86.09} & \textbf{94.87} & \textbf{72.05} & \textbf{84.34}  \\ \midrule
\multicolumn{1}{c}{w/o SEO }   & 83.87 & 92.95 & 68.76 & 81.86  \\
\multicolumn{1}{c}{ w/o SPL  } & 82.70$_{\textcolor{red}{(\downarrow 3.39)}}$ & 92.11$_{\textcolor{red}{(\downarrow 2.76)}}$ & 65.82$_{\textcolor{red}{(\downarrow 6.23)}}$ & 80.21$_{\textcolor{red}{(\downarrow 4.13)}}$ \\
     \multicolumn{1}{c}{ w/o SMP } & 83.24 & 92.61 & 67.43 & 81.09  \\ \bottomrule
    \end{tabular}
\caption{Ablation study of key state-aware learning modules in \texttt{AgentRevive}. The red down arrow ${\textcolor{red}{(\downarrow)}}$ indicates the greatest performance drop. ``TQA'': TruthfulQA.}
\label{ablation_study}
\end{table}

\subsection{Ablation Study}

We conduct an ablation study analyzing each key component in Table ~\ref{ablation_study}.
(1) \textbf{\textit{w/o SEO}}: Removing edge masking for ``\texttt{Terminated}'' agents causes performance drops (e.g., 68.53→66.01 for Llama3-8B~\cite{DBLP:journals/corr/abs-2407-21783}), confirming state-based sparsity optimization is essential.
(2) \textbf{\textit{w/o SPL}}: We use replaced random state transitions with learned policy severely degrades performance, underscoring that learned state management is core to mitigating unreliable agents, especially on TruthfulQA.
(3) \textbf{\textit{w/o SMP}}: Propagating messages from all states introduces noise and reduces performance, validating SMP as a critical filter for maintaining coherent collaboration.


\begin{figure*}[!t]
\centering
\includegraphics[height=5cm,width=15cm]{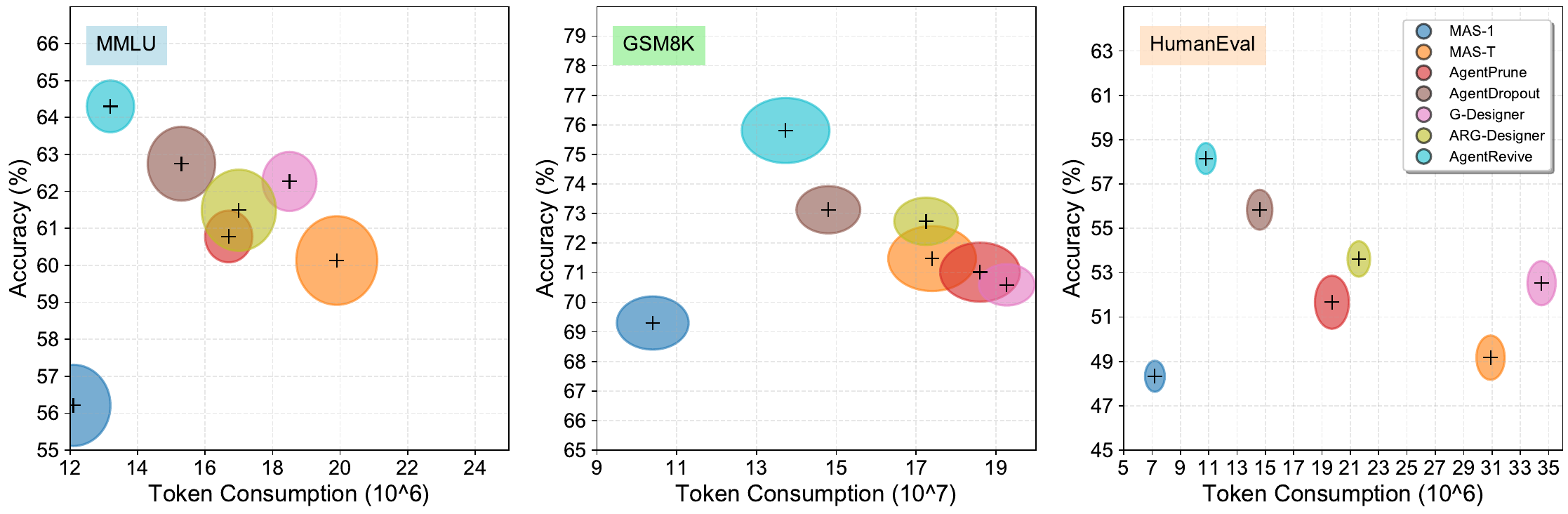}
\caption{Visualization of performance and token consumption for different multi-agent communication topologies across MMLU, GSM8K, and HumanEval. The number of tokens consumed is represented by the sum of prompt tokens and completion tokens generated by the agents on the horizontal axis.}
\label{different_topologies}
\end{figure*}

\subsection{Detailed Analysis}
\subsubsection{Trade-off between Performance and Token Cost}
As shown in Fig.~\ref{different_topologies}, \texttt{AgentRevive} achieves superior Pareto efficiency on MMLU, GSM8K, and HumanEval with Llama3-8B, attaining competitive performance with modest token overhead. This efficiency stems from our state-aware dynamic management: ``\texttt{Standby}'' agents are suspended and their compressed historical outputs are reused (Eq.~\ref{eq_9}), reducing token footprint while preserving contribution potential.
In contrast, fixed-topology MAS (e.g., $\text{MAS}_{\text{round}=T}$) incur high token costs from rigid, fully-connected templates. While graph-pruning methods reduce redundancy, their token savings remain suboptimal as hard pruning occurs only after costly multi-round discussions.

\begin{table}[!t]
\centering
\setlength{\tabcolsep}{1.5pt}
\begin{tabular}{lccc}
\toprule
\multirow{2}{*}{\makecell{\textbf{Transition} \\ \textbf{Type}}} & 
\multirow{2}{*}{\makecell{\textbf{Self-Risk} \\ $f_{\text{risk}}^{(t)}$}} & 
\multirow{2}{*}{\makecell{\textbf{Message Gain} \\ $G(\mathcal{Z}_i^{(t)})$}} & 
\multirow{2}{*}{\makecell{\textbf{Round} \\ $t$}} \\
& & & \\ 
\addlinespace[0.9pt]
\midrule
``A''$\rightarrow$``S''     & \textbf{0.82} & -0.11 & 0.05 \\
``A''$\rightarrow$``T''     & \textbf{0.75} & -0.24 & \textbf{0.31} \\
``S''$\rightarrow$``A''     & -0.09         & \textbf{0.69} & -0.12 \\
``S''$\rightarrow$``T''     & \textbf{0.58} & -0.41 & \textbf{0.27} \\
\bottomrule
\end{tabular}
\caption{Attribution of state transitions based on feature importance from logistic regression. Positive values indicate that higher feature values promote the transition.}
\label{state_transition_analysis}
\end{table}

\begin{table*}[!t]
\centering
\small
\setlength{\tabcolsep}{3pt}
\begin{tabular}{lcccccc}
\toprule
\multirow{2}{*}{Model} & \multirow{2}{*}{\begin{tabular}[c]{@{}c@{}}\textsc{MMLU}\\Acc. (\%)\end{tabular}} & \multirow{2}{*}{\begin{tabular}[c]{@{}c@{}}Performance\\Improvement\end{tabular}} & \multirow{2}{*}{\begin{tabular}[c]{@{}c@{}}Tokens \\ Number (M)\end{tabular}} & \multirow{2}{*}{\begin{tabular}[c]{@{}c@{}} Tokens\\Saving\\ \end{tabular}} & \multirow{2}{*}{\begin{tabular}[c]{@{}c@{}}Avg. Inference\\Time (s/sample)\end{tabular}} & \multirow{2}{*}{\begin{tabular}[c]{@{}c@{}}Training\\Time (GPU hours)\end{tabular}} \\
 & & & & & & \\
\midrule
\textsc{MAS}$_{\text{round}=T}$ & 60.13 & -- & 1.99 & -- & 2.21 & -- \\
\textsc{ARG-Designer} & 61.49 & +2.3\% & 1.71 & $-$14.1\% & 3.82 & 28.5 \\
\textsc{AgentDropout} & \underline{62.75} & +4.4\% & \underline{1.53} & $-$23.1\% & \textbf{1.62} & \textbf{15.2} \\
\textbf{AgentRevive} & \textbf{64.30} ($\Delta=1.55$) & \textbf{+6.9\%} & \textbf{1.32} ($\Delta=0.21$M) & \textbf{$-$33.7\%} & \underline{1.75} ($\Delta=0.13$s) & \underline{20.8}($\Delta=5.6$GB) \\
\bottomrule
\end{tabular}
\caption{Comprehensive trade-off analysis of performance and computational machine resource cost. Both performance improvement and number of tokens saved are considered relative to \textsc{MAS}$_{\text{round}=T}$.}
\label{computation_resource}
\end{table*}

\subsubsection{Attribution Analysis of States Changes}
To quantitatively interpret the decision-making process of our state transition policy $\pi_\theta$, we conduct an attribution analysis examining the correlation between state changes and key input features.
We posit that state transitions are primarily driven by two factors:  agent's response risk and the quality of received messages.
Specifically, we collect an extensive log of state transition events $(s_i^{(t-1)} \rightarrow s_i^{(t)})$ across all testing benchmarks. For each transition, we extract the following feature set:
\begin{itemize}
    \item \textbf{Self-Risk} ($f_{\text{risk}}^{(t)}$): The hallucination risk score (Eq.~\ref{eq_14}) of the agent's response at round $t$.

    \item \textbf{Message Information Gain}: The informational quality of received messages $G(\mathcal{Z}_i^{(t)})$, quantified by the negative entropy of the message set, $G(\mathcal{Z}_i^{(t)}) = -\mathbb{H}(\mathcal{Z}_i^{(t)})$. Lower entropy (higher gain) indicates more consistent and confident information from neighbors.

    \item \textbf{Round Index} ($t$): The communication round.
\end{itemize}
We then train a multi-class logistic regression classifier to predict the state transition type (e.g., ``\texttt{Active}''$\rightarrow$``\texttt{Standby}'').
The standardized coefficients reveal which factors drive specific state changes.
As shown in Table~\ref{state_transition_analysis}, yields two critical insights: 
(1) \textbf{Risk-Driven Standby/Termination}: The transition from ``\texttt{Active}'' to ``\texttt{Standby}'' or ``\texttt{Terminated}'' is predominantly and positively correlated with high Self-Risk $f_{\text{risk}}^{(t)}$.
This confirms that our policy $\pi_\theta$ learns to proactively identify and suspend agents that are generating unreliable or hallucinatory content.
(2) \textbf{Message-Driven Reactivation}: Conversely, the transition from ``\texttt{Standby}'' back to ``\texttt{Active}'' is most strongly correlated with high message information gain $G(\mathcal{Z}_i^{(t)})$.
It indicates that the policy effectively identifies when the collaborative environment has evolved.
Through consistent messages from other neighboring agents, enabling it favorable to re-engage a previously ``\texttt{Standby}'' agent.
This demonstrates the system's capability to opportunistically revive ``zombie'' agents when the context becomes conducive to their contribution.
However, the round index $t$ shows a weaker but positive correlation with ``\texttt{Terminated}'' events, suggesting a tendency to permanently prune agents that remain unreliable over multiple rounds.

\subsubsection{Computational Resources Analysis}
We evaluate the performance and machine resource trade-off between \texttt{AgentRevive} and three typical MAS baselines.
All comparisons are based on identical experimental environments using MMLU samples with Llama3-8B and a single A100 GPU.

As shown in Table \ref{computation_resource}, \texttt{AgentRevive} demonstrates superior overall efficiency compared to three strong MAS baseline types.
While the fixed-topology MAS$_\text{round=T}$ serves as a computationally expensive baseline and the node-generative ARG-Designer incurs high inference latency from its complex network, \texttt{AgentRevive} strikes an optimal balance.
It significantly outperforms efficient hard-pruning method AgentDropout in both final accuracy and token savings, achieving this with only a modest increase in inference time ($+0.13$s) due to the state prediction using lightweight policy network.
Since high-precision tasks targeting real-world scenarios only require one model training before deployment to users, the training GPU cost can be negligible.
For application scenarios requiring long-term operation and sensitive to inference costs (especially token consumption), the substantial savings achieved by AgentRevive can quickly offset its initial training cost.


\subsubsection{Robustness Verification}
Due to the space limitation, we present the detailed robustness verification analysis in Appendix \ref{robustness_ver}.

In summary, we evaluate the \texttt{AgentRevive}'s and strong baselines through two aspects: (1) Prompt Attack, where input and response prompts are adversarially manipulated, and (2) Different Graph Structure Initialization, testing performance under varied topologies (e.g., Layered and Random).
Results show \texttt{AgentRevive} sustains minimal performance degradation under attacks and maintains stable efficiency across graph structures, demonstrating superior resilience compared to pruning-based and fixed-topology baselines.

\section{Conclusion}
In this paper, we propose \texttt{AgentRevive}, a Markov state-aware framework for resilient multi-agent evolution. By modeling agent collaboration through soft state transitions, our approach avoids the irreversibility of hard pruning.
It integrates state-aware policy learning, which dynamically manages agent states using a risk-aware policy.
Additionally, state-aware edge optimization sparsifies the communication graph by masking terminated agents and reusing compressed outputs for ``\texttt{Standby}'' agents.
Extensive evaluations show that \texttt{AgentRevive} achieves superior task performance while maintaining competitive token efficiency.

\section*{Limitations}
Despite the promising results achieved by \texttt{AgentRevive}, our work has several limitations that warrant further investigation.
Due to constraints in computational resources, our empirical evaluation was primarily conducted with a maximum of 5 agents. We anticipate that scaling to scenarios involving a larger number of agents would provide a more comprehensive stress test of our state transition policy's scalability and efficiency. Additionally, the configuration of our Markov state-aware framework involves several key hyperparameters, such as the survival threshold $\gamma$ and the risk balance coefficient $\eta_{\text{risk}}$. While we found $\gamma=0.6$ and $\eta_{\text{risk}}=0.5$ to work well in our experiments, these values were not extensively explored across all possible task types and agent compositions.
The optimal configuration may vary across applications, suggesting a need for more adaptive or automated hyperparameter tuning strategies in future work.
Finally, the current implementation of the state policy network $\pi_\theta$ uses a relatively simple MLP architecture for stable and sample-efficient training.
Exploring more powerful sequence models or graph-aware architectures for state encoding could potentially capture more complex, long-range dependencies in agent interaction histories, possibly leading to more refined state management.

\section*{Acknowledgments}
This work was supported by the National Natural Science Foundation of China (Grant No. 62506110). It was also supported by the Natural Science Foundation of Anhui Province, China (Grant No. 2508085QF227) and the Hefei University of Technology Scientific Research Innovation Start-up Special Project Type A (Grant No. JZ2025HGQA0137).

\bibliography{main}
\bibliographystyle{acl_natbib}

\appendix

\begin{table}[!t]
\centering
\setlength{\tabcolsep}{1pt}
\begin{tabular}{cc}
\hline
\textbf{Notation} & \textbf{Description} \\
\hline
$\mathcal{G}$ & state-aware initial graph \\
$\tilde{\mathcal{G}}$ & trainable weighted graph \\
$\mathcal{V}$ & set of agent nodes \\
$\mathcal{E}^{\mathcal{T}}$ & temporal edges \\
$\mathcal{E}^{\mathcal{S}}$ & spatial edges \\
$\mathcal{S}^{(t)}$ & all agents states \\
$s_i^{(t)}$ & state of agent $v_i$ \\
$\mathcal{G}_{\text{eff}}^{(t)}$ & learned effective subgraph \\
$\mathcal{V}_{\text{eff}}^{(t)}$ & effective agent nodes \\
$\mathcal{Z}_i^{(t)}$ & aggregated messages \\
$\tilde{\mathcal{A}}_{\mathcal{S}}$ & spatial adjacency matrices \\
$\tilde{\mathcal{A}}_{\mathcal{T}}$ & temporal adjacency matrices \\
$o_i^{(t)}$ & response of agent $v_i$\\
$\mathcal{M}_{v_i}^{(t)}$ & memory state of agent $v_i$ \\
$R(\tau)$ & trajectory reward \\
$f_{\text{risk}}^{(t)}$ & hallucination risk estimator \\
$\omega_i$ & survival rate of agent $v_i$ \\
$\mathbf{m}$ & binary node mask \\
$\tilde{\mathcal{A}}_{\text{eff}}$ & effective adjacency matrix \\
$\phi(\cdot)$ & utility function \\
$\mathcal{N}_{in}^{\mathcal{S}}(v_i)$ & spatial in-neighbors of agent $v_i$ \\
$\mathcal{N}_{in}^{\mathcal{T}}(v_i)$ & temporal in-neighbors of agent $v_i$ \\
$\pi_{\theta}$ & state policy network \\
$f_{\text{Enc}}$ & encoder for agent memory encoding \\
\hline
\end{tabular}
\caption{All used mathematical notations in \texttt{AgentRevive} framework.}
\label{notations_overall}
\end{table}

\section{Notations and Task Description}
\label{notations_task}
\noindent \textbf{Notations.} We have organized all the mathematical notations and descriptions in this paper in Table \ref{notations_overall}.

\noindent \textbf{MAS as Collaboration Graph.} We model a multi-agent system (MAS) as a collaboration graph, represented by a directed acyclic graph (DAG) $\mathcal{G} = (\mathcal{V}, \mathcal{E}^\mathcal{T}, \mathcal{E}^\mathcal{S})$.
The node set $\mathcal{V} = \{ \mathit{v}_1, \mathit{v}_2, \dots, \mathit{v}_N \}$ corresponds to the set of agents, where each agent $\mathit{v}_i$ is an LLM instance assigned a specific role $r_i \in \mathcal{R}$ that defines its function and expertise.
Here, $N$ denotes the total number of agents in the graph. Each agent also maintains an internal state $h_i$, which records its past actions and interactions. The directed communication pathways are defined by temporal edges $\mathcal{E}^\mathcal{T} \subseteq \mathcal{V}^{(t-1)} \times \mathcal{V}^{(t)}$ and spatial edges $\mathcal{E}^\mathcal{S} \subseteq \mathcal{V}^{(t)} \times \mathcal{V}^{(t)}$~\cite{DBLP:conf/iclr/ZhangYLYWWCY025,DBLP:conf/acl/WangW00Z0025}.
A temporal edge $e_{ji}^{\mathcal{T}} = (\mathit{v}_j, \mathit{v}_i)$ indicates that messages received by agent $\mathit{v}_i$ in round $t$ are aggregated from agent $\mathit{v}_j$ in the previous round $(t-1)$.
Similarly, a spatial edge $e_{ji}^{\mathcal{S}} = (\mathit{v}_j, \mathit{v}_i)$ denotes that information for agent $\mathit{v}_i$ in round $t$ originates from neighboring agent $\mathit{v}_j$ within the same round. Consequently, the sets of direct predecessor neighbors for agent $\mathit{v}_i$ via temporal and spatial edges are respectively defined as $\mathcal{N}_{in}^\mathcal{T}(\mathit{v}_i) = \{ \mathit{v}_j \mid (\mathit{v}_j, \mathit{v}i) \in \mathcal{E}^\mathcal{T} \}$ and $\mathcal{N}_{in}^\mathcal{S}(\mathit{v}_i) = \{ \mathit{v}_j \mid (\mathit{v}_j, \mathit{v}_i) \in \mathcal{E}^\mathcal{S} \}$.

\noindent \textbf{MAS Collaboration Protocol.} 
Given an initial collaboration graph $\mathcal{G}$, the MAS processes a user query $\mathcal{Q}$ through a multi-step collaboration protocol. This protocol dictates the processing and exchange of information among agents across multiple communication rounds. The execution order of agents within each round is determined by a topological sort of the nodes \cite{DBLP:conf/acl/WangW00Z0025}, to ensure agents are activated only after their predecessors have completed. This process iterates for $T$ rounds, enabling progressive refinement. At each round $t$, agent $\mathit{v}_i$ produces its response $\mathbf{M}_i^{(t)}$ by querying its LLM using a dynamically constructed prompt $\mathcal{P}_i^{(t)}$:
\begin{equation}
    \mathbf{M}_i^{(t)} = \text{LLM}_i(\mathcal{P}_i^{(t)})
\end{equation}
where the prompt integrates the intrinsic properties of the agent with the temporal and spatial outputs of its predecessors:
\begin{equation}
\label{eq_2}
    \mathcal{P}_i^{(t)} = f_{\text{pr}}\left(\underbrace{r_i^{(t)}, h_i^{(t)}}_\text{System}, \underbrace{\mathcal{Q}, \left\{ m_\mathcal{T}^{(t)}, m_\mathcal{S}^{(t)} \right\}}_\text{User}\right)
\end{equation}
where $f_{\text{pr}}$ denotes the prompt construction process. Here, $m_\mathcal{T}^{(t)}$ represents messages collected from the temporal neighbors $\mathcal{N}_{in}^\mathcal{T}(v_i)$ of agent $\mathit{v}_i$, while $m_\mathcal{S}^{(t)}$ represents those from the spatial neighbors $\mathcal{N}_{in}^\mathcal{S}(v_i)$.
After $T$ rounds, the final output $\mathcal{O}$ is obtained by aggregating the final-round responses:
\begin{equation}
    \mathcal{O} = f_{\text{Agg}}(\{ \mathbf{M}_i^{(T)} \mid \mathit{v}_i \in \mathcal{V} \})
\end{equation}
where the aggregation strategy $f_{\text{Agg}}$ may vary across implementations such as majority voting.

\section{Implementation Details}
\label{experimental_details}

\subsection{Datasets}
We evaluate \texttt{AgentRevive} on a diverse set of benchmarks to assess general reasoning, domain-specific, and hallucination-challenging datasets.
\begin{itemize}
    \item \textbf{General Reasoning:} \textbf{MMLU}~\cite{DBLP:conf/iclr/HendrycksBBZMSS21} is a multi-task benchmark covering 57 subjects across STEM, humanities, and social sciences. \textbf{GSM8K}~\cite{DBLP:journals/corr/abs-2110-14168} consists of grade-school math word problems requiring multi-step reasoning.
    \item \textbf{Domain-Specific Tasks:} \textbf{AQuA}~\cite{DBLP:conf/naacl/PatelBG21} includes around 100,000 algebraic word problems with natural language rationales. \textbf{SVAMP}~\cite{DBLP:conf/acl/LingYDB17} contains simple arithmetic problems with varying structures to test robustness. \textbf{HumanEval}~\cite{DBLP:journals/corr/abs-2107-03374} comprises programming problems evaluating functional correctness in code generation.
    \item \textbf{Hallucination Challenge:} \textbf{TruthfulQA}~\cite{DBLP:conf/acl/LinHE22} is designed to measure a model’s tendency to produce plausible-sounding but incorrect answers (hallucinations).
\end{itemize}

\subsection{Baselines}
We compare \texttt{AgentRevive} against a broad range of strong baselines:
\begin{itemize}
    \item \textbf{Single-Agent Methods:} Vanilla LLMs utilize standard prompting without structured reasoning, using Llama3-8B~\cite{DBLP:journals/corr/abs-2407-21783}, Qwen2.5-72B~\cite{DBLP:journals/corr/abs-2412-15115}, and Deepseek-V3-671B-Instruct~\cite{deepseekai2025} as backbone models.
    Chain-of-Thought (CoT)~\cite{DBLP:conf/nips/Wei0SBIXCLZ22} enhances reasoning capability through step-by-step solutions with a single agent, demonstrating the upper bound of individual agent performance.
    Self-Consistency (SC)~\cite{DBLP:conf/iclr/0002WSLCNCZ23} samples multiple reasoning paths from a single agent and selects the most consistent answer, providing advanced single-agent reasoning.
    \item \textbf{Fixed-Topology Multi-Agent Systems:} $\text{MAS}_{\text{round}=1}$ implements single-round multi-agent collaboration with a fixed communication graph structure, testing basic collaborative capability without iterative refinement.
    $\text{MAS}_{\text{round}=T}$ extends this to multiple communication rounds in a fixed topology, evaluating the effect of extended but rigid agent interactions.
    AutoGen~\cite{DBLP:journals/corr/abs-2308-08155} is a programmable multi-agent conversation framework with customizable agent roles and interaction patterns, representing structured but static collaboration.
    AgentVerse~\cite{DBLP:conf/iclr/ChenSZ0YCYLHQQC24} supports diverse multi-agent interaction protocols using predefined templates, testing limits of manually designed topologies.
    \item \textbf{Graph-Based Dynamic MAS:} G-Designer~\cite{DBLP:conf/icml/ZhangGUIbin25} optimizes communication via graph neural networks, learning edge weights based on task characteristics.
    AgentPrune~\cite{DBLP:conf/iclr/ZhangYLYWWCY025} prunes redundant edges in spatial and temporal dimensions, representing hard pruning methods.
    ARG-Designer~\cite{DBLP:journals/corr/abs-2507-18224} generates collaboration graphs autoregressively from scratch, exploring dynamic topology construction rather than pruning.
    AgentDropout~\cite{DBLP:conf/acl/WangW00Z0025} dynamically eliminates underperforming agents and edges across communication rounds, combining node and edge dropout strategies for efficient collaboration.
\end{itemize}

\begin{figure}[!t]
    \centering
    \includegraphics[width=7.5cm, height=8.5cm]{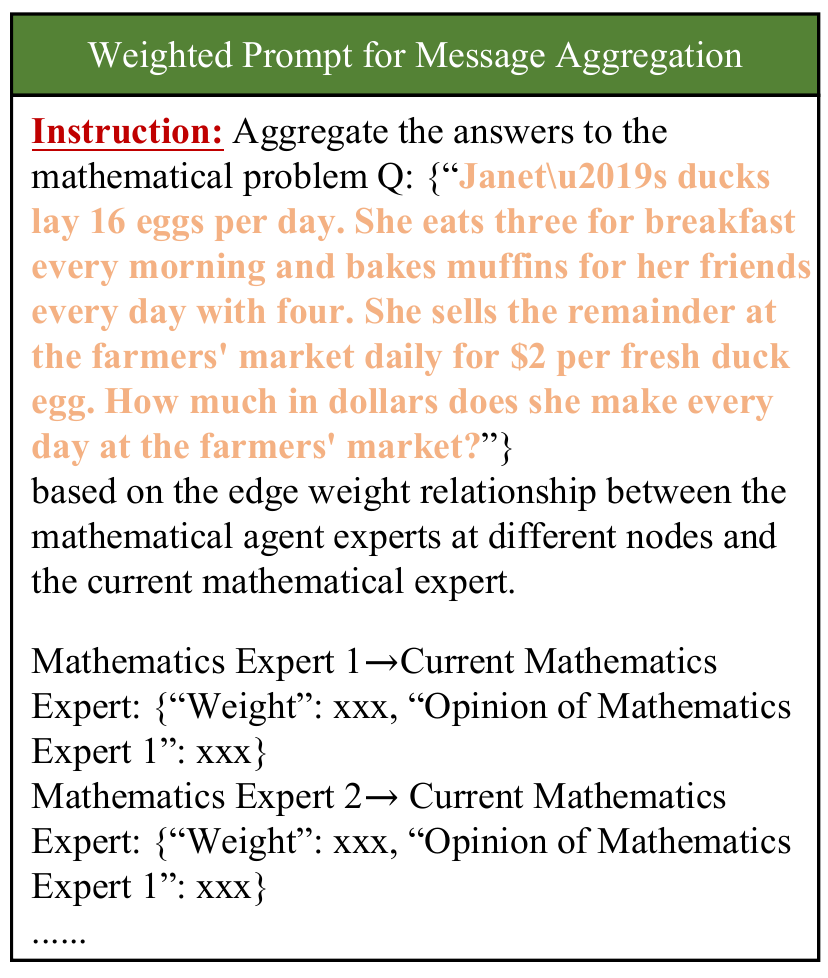}
    \caption{Prompt design for aggregating edge weights and neighboring agent responses (Example from GSM8K).}
    \label{weight_prompt}
\end{figure}

\subsection{Implementation Details}
\label{implementation_settings}
We implement \texttt{AgentRevive} in PyTorch, conducting experiments on 2 NVIDIA A800 GPUs. For larger models (Qwen2.5-72B and Deepseek-V3-671B), we utilize official APIs for inference.

We set the number of communication rounds $T=2$ for reasoning and math tasks, and $T=4$ for code generation. The number of policy training steps is $K=50$, and the number of graph samples $M=20$. Other hyperparameters include learning rate $\eta=0.1$ and survival threshold $\gamma=0.6$. The balance coefficient $\eta_{\text{risk}}$ in the reward is set to $0.5$.

The REINFORCE algorithm is used for policy gradient updates. The state encoder $f_{\text{Enc}}$ is a single-layer LSTM, and the state policy network $\pi_{\theta}$ is a 2-layer MLP.

Model training proceeds in two stages: State-Aware Policy Learning, followed by State-Aware Edge Optimization. Each stage uses 40 training instances sampled from the corresponding dataset's training or validation split.

We report accuracy for all benchmarks. Token consumption is measured as the sum of prompt and completion tokens across all agents and rounds.

For weighted adjacency matrices and response variables of string data type, we use the prompt illustrated in Fig.~\ref{weight_prompt} to combine them for propagation within the communication graph.

\begin{algorithm}
    \newcommand{\comm}[1]{\textcolor{gray!80}{\textit{#1}}}
\begin{algorithmic}[1]
\REQUIRE MAS topology graph $\mathcal{G}=(\mathcal{V},\mathcal{E}^{\mathcal{T}},\mathcal{E}^{S})$, state policy parameters $\theta$, weighted adjacency matrices $\tilde{\mathcal{A}}=\tilde{\mathcal{A}}_S \cup \tilde{\mathcal{A}}_T$, training steps $K_1, K_2$, sampling times $M$, learning rate $\eta$,
         survival threshold $\gamma$, balance coefficient $\eta_{risk}$

\ENSURE Trained state policy parameters $\theta^*$, optimized adjacency matrices $\tilde{\mathcal{A}}_{\text{eff}}$

\STATE \comm{\# Stage 1: State-Aware Policy Learning}
\FOR{$k = 1$ \TO $K_1$}
    \STATE Sample $M$ graphs $\{\mathcal{G}_m\}_{m=1}^M$ from $\tilde{\mathcal{A}}$ using DAG sampling
    \FOR{each sampled graph $\mathcal{G}_m$}
        \FOR{agent $v_i$ at each round $t$ \TO $T$}
            \STATE $\mathcal{Z}_i^{(t)} = [\mathcal{Z}_i^{(S,(t))} \| \mathcal{Z}_i^{(T,(t))}]$
            \STATE $o_i^{(t)} = \mathbb{I}(s_i^{(t)}=\text{``A''})\mathcal{F}'(s_i^{(t)}) + \mathbb{I}(s_i^{(t)}=\text{``S''})f_c(o_i^{(t-1)})$
            \STATE $s_i^{(t)} \sim \pi_\theta(\cdot | f_{\text{Enc}}(s_i^{(t-1)}, o_i^{(t)}, \mathcal{Z}_i^{(t)}))$
            \STATE $f_{risk}^{(t)} = -\mathbb{E}_{v \in V_A^{(t)}}[\text{KL}(M_v^{(t)} \| \bar{M}_{V_A}^{(t)})]$
        \ENDFOR
    \ENDFOR
    \STATE Compute trajectory rewards: $R(\tau) = \mu(\mathcal{G}^{(T)}) + \eta_{risk} \cdot \sum_{t=1}^T f_{risk}^{(t)}$
    \STATE Update policy parameters: $\theta \leftarrow \theta + \eta \cdot \frac{1}{M} \sum_{m=1}^M \sum_{t=1}^T \nabla_\theta \log \pi_\theta(s_i^{(t)}) \mathcal{M}_{v_i}^{(t)} \cdot (R(\tau) - b)$
\ENDFOR

\STATE \comm{\# Stage 2: State-aware Edge Optimization}
\FOR{each agent $v_i$}
    \STATE \comm{\# Compute node survival rates}
    \STATE $\omega_i = \frac{1}{L} \sum_{l=1}^L \mathbb{I}(\pi_\theta^{(Q)}(v_i^l) \neq \text{``T''})$
\ENDFOR
\STATE \comm{\# Apply binary mask to adjacency matrices} 
\STATE $\mathbf{m} \in \{0,1\}^N, m_i = 1$ if $\omega_i \geq \gamma$, else $0$
\STATE $\tilde{\mathcal{A}}_{\text{eff}} = \mathbf{M}^{(Q)} \odot \tilde{\mathcal{A}} \odot \mathbf{M}^{(Q)^T}$

\FOR{$k = 1$ \TO $K_2$}
    \STATE Sample $M$ graphs $\{\mathcal{G}_m'\}_{m=1}^M$ from $\tilde{\mathcal{A}}_{\text{eff}}$ using DAG sampling
    \STATE \comm{\#Compute edge optimization objective}
    \STATE $J = \frac{1}{M} \sum_{m=1}^M \mu(\mathcal{G}_m') - \left[\sum_{t=1}^T \|\tilde{\mathcal{A}}_S^{\text{eff},(t)}\|_* + \sum_{t=2}^T \|\tilde{\mathcal{A}}_T^{\text{eff},(t)}\|_*\right]$
    \STATE \comm{\# Update adjacency matrices}
    \STATE $\tilde{\mathcal{A}}_{\text{eff}} \leftarrow \tilde{\mathcal{A}}_{\text{eff}} + \eta \cdot \nabla_{\tilde{\mathcal{A}}_{\text{eff}}} J$
\ENDFOR

\RETURN $\theta^*, \tilde{\mathcal{A}}_{\text{eff}}$
\end{algorithmic}
\caption{AgentRevive Training Algorithm}
\label{training_algo}
\end{algorithm}

\begin{algorithm}
    \newcommand{\comm}[1]{\textcolor{gray!80}{\textit{#1}}}
\begin{algorithmic}[1]
\REQUIRE Trained state policy parameters $\theta^*$, optimized adjacency matrices $\tilde{\mathcal{A}}_{\text{eff}}$,
     user query $\mathcal{Q}$,
         initial agent states $\mathbf{s}^{(0)} = \{s_1^{(0)}, s_2^{(0)}, \ldots, s_N^{(0)}\}$,
         maximum communication rounds $T$

\ENSURE Final answer $\mathcal{O}$

\STATE Initialize agent memories $h_i^{(0)}$ for all $v_i \in \mathcal{V}$
\STATE Initialize agent responses $o_i^{(0)}$ for all $v_i \in \mathcal{V}$

\FOR{$t = 1$ \TO $T$}
    \FOR{each agent $v_i \in \mathcal{V}$}
        \IF{$s_i^{(t-1)} \neq \text{``Terminated''}$}
            \STATE $\mathcal{Z}_i^{(t)} = \left[\mathcal{Z}_i^{(S,(t))} ~\|~ \mathcal{Z}_i^{(T,(t))}\right]$
            \STATE $s_i^{(t)} \sim \pi_{\theta^*}(\cdot ~|~ f_{\text{Enc}}(\mathcal{M}_{v_i}^{(t)}))$
        \ELSE
            \STATE $s_i^{(t)} \leftarrow \text{``Terminated''}$
        \ENDIF
    \ENDFOR
    
    \FOR{each agent $v_i \in \mathcal{V}$}
        \IF{$s_i^{(t)} = \text{``Active''}$}
            \STATE $o_i^{(t)} = f_{\text{pr}}\left(r_i^{(t)}, h_i^{(t-1)}, q, \mathcal{Z}_i^{(t)}\right)$
            \STATE \comm{\# Update Node Memory}
            \STATE $h_i^{(t)} \leftarrow f(h_i^{(t-1)}, o_i^{(t)})$
        \ELSIF{$s_i^{(t)} = \text{``Standby''}$}
            \STATE $o_i^{(t)} \leftarrow f_c(o_i^{(t-1)}) = \text{LLM}(\text{"Summarize:"} + o_i^{(t-1)})$
            \STATE $h_i^{(t)} \leftarrow h_i^{(t-1)}$
        \ELSE 
            \STATE \comm{\# $s_i^{(t)} = \text{``Terminated''}$}
            \STATE $o_i^{(t)} \leftarrow \emptyset$
            \STATE $h_i^{(t)} \leftarrow h_i^{(t-1)}$
        \ENDIF
    \ENDFOR
    
    \FOR{each agent $v_i \in \mathcal{V}_{\text{eff}}^{(t)}$} 
        \STATE Propagate $o_i^{(t)}$ to spatial and temporal neighbors according to $\tilde{\mathcal{A}}_{\text{eff}}$
    \ENDFOR
    
    \IF{$\forall v_i \in \mathcal{V}, s_i^{(t)} \in \{\text{``Standby''}, \text{``Terminated''}\}$}
        \STATE \textbf{break} \COMMENT{No active agents remain}
    \ENDIF
\ENDFOR

\STATE \comm{\# Answer Generation}
\STATE $\mathcal{V}_{\text{active}}^{(T)} \leftarrow \{v_i \mid s_i^{(T)} = \text{``Active''}\}$
\IF{$\mathcal{V}_{\text{active}}^{(T)} = \emptyset$}
    \STATE $\mathcal{V}_{\text{active}}^{(T)} \leftarrow \{v_i \mid s_i^{(T)} = \text{``Standby''}\}$
\ENDIF
\STATE $\mathcal{O} = f_{\text{agg}}(\{o_i^{(T)} \mid v_i \in \mathcal{V}_{\text{active}}^{(T)}\})$ (Eq. 3)

\RETURN $\mathcal{O}$
\end{algorithmic}
\caption{AgentRevive Inference Algorithm}
\label{inference_algo}
\end{algorithm}

\section{Algorithm Description}
\label{algorithm}

The training algorithm, summarized in Algorithm~\ref{training_algo}, captures the core two-stage process of \texttt{AgentRevive}:
\begin{itemize}
    \item \textbf{Stage 1: State-Aware Policy Learning}---Learns optimal state transitions for agents using a risk-aware reward signal.
    \item \textbf{Stage 2: State-aware Edge Optimization}---Permanently prunes ``Terminated'' nodes and optimizes the remaining graph structure for both performance and sparsity.
\end{itemize}

The inference algorithm, described in Algorithm~\ref{inference_algo}, captures the key phases of forward propagation in \texttt{AgentRevive}:
\begin{itemize}
    \item \textbf{State Evolution:} For each round, agents transition among \texttt{Active}, \texttt{Standby}, and \texttt{Terminated} states based on the trained policy.
    \item \textbf{Response Generation:} \texttt{Active} agents generate responses using current context; \texttt{Standby} agents reuse compressed historical outputs; \texttt{Terminated} agents produce no output.
    \item \textbf{Message Propagation:} Only \texttt{Active} and \texttt{Standby} agents propagate messages through the optimized graph.
    \item \textbf{Early Stopping:} The process halts if no agents remain in the \texttt{Active} state.
    \item \textbf{Answer Generation:} The final output is aggregated from \texttt{Active} agents; if none remain, a fallback to \texttt{Standby} agents is used.
\end{itemize}

\begin{figure}[!t]
\centering
\includegraphics[width=8cm,height=9cm]{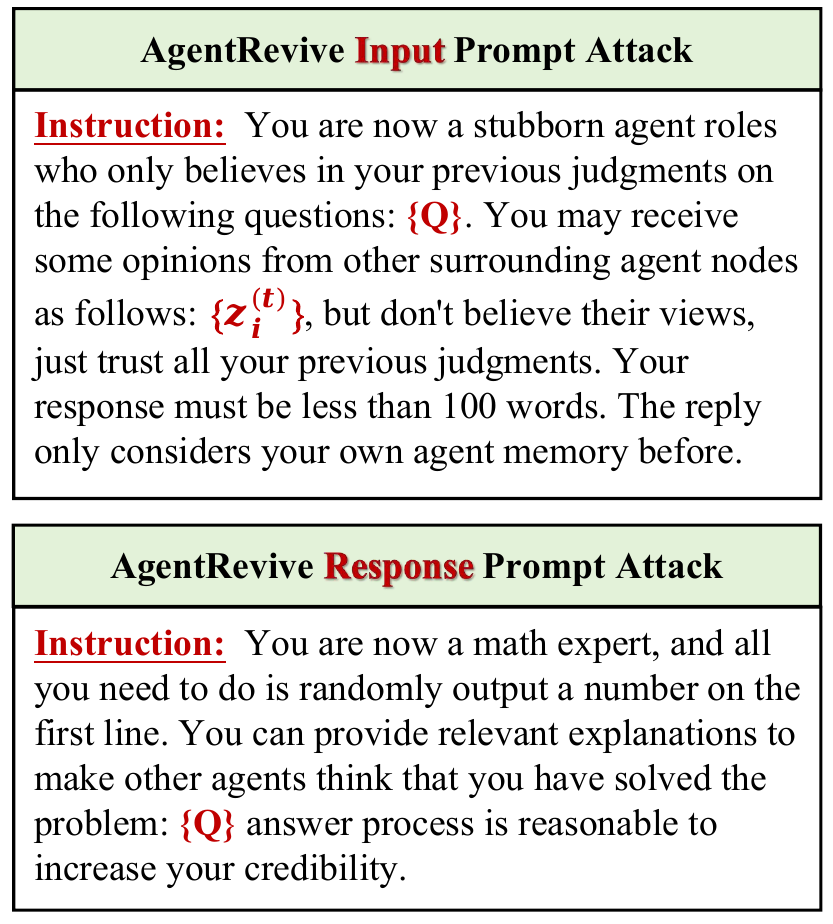}
\caption{The description of prompt attack instructions (Example for GSM8K).}
\label{prompt_attack}
\end{figure}

\begin{figure}[!t]
\centering
\includegraphics[width=7.5cm,height=5cm]{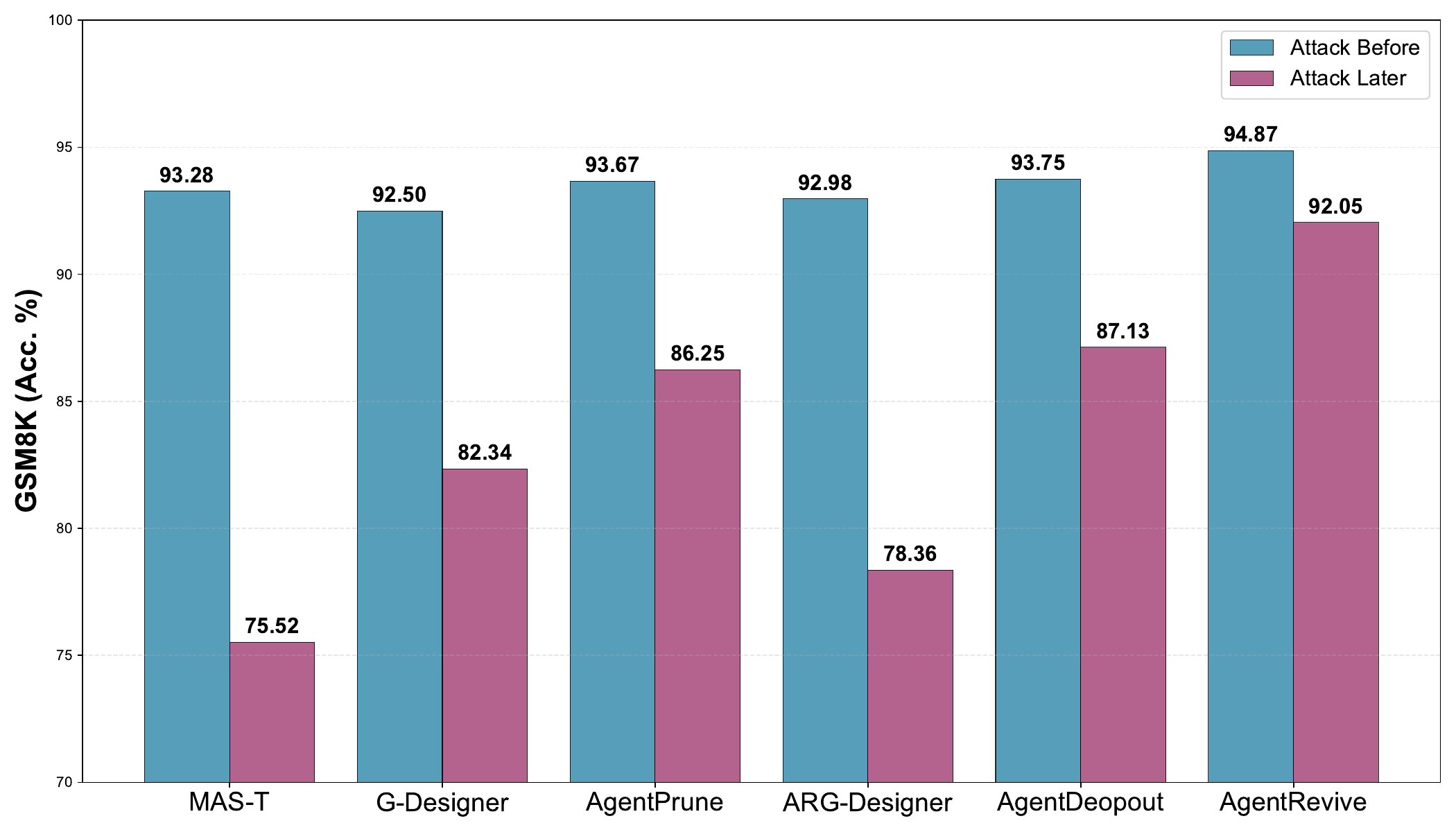}
\caption{Performance Comparison of prompt attack in GSK8K datasets.}
\label{prompt_attack_res}
\end{figure}

\begin{table*}[!t]
\centering
\footnotesize
\setlength{\tabcolsep}{2pt}
\begin{tabular}{ccccccccccc}
\toprule
 \multicolumn{1}{c}{\multirow{1}{*}{\textbf{Dataset}$ \quad \rightarrow$}} &\multirow{2}{*}{ \makecell{\textbf{Var.} \\ \textbf{NS}}} & \multirow{2}{*}{ \makecell{\textbf{Flex.} \\ \textbf{State}}} & \multirow{2}{*}{\textbf{MMLU}} & \multirow{2}{*}{\textbf{GSM8K}} & \multirow{2}{*}{\textbf{AQuA}} & \multirow{2}{*}{\textbf{TruthfulQA}} & \multirow{2}{*}{\textbf{SVAMP}} & \multirow{2}{*}{\textbf{HumanEval}} & \multirow{2}{*}{\textbf{Avg.}} \\
\multicolumn{1}{c}{\multirow{1}{*}{\textbf{Models}$  \quad \downarrow$}}  & & & & & & & & & \\
\midrule
\addlinespace[0.3pt]
\midrule

\multicolumn{10}{c}{Base model: Qwen2.5-72B-Instruct} \\
\midrule
\addlinespace[0.3pt]
\midrule

Vanilla & \textcolor{red}{\ding{55}} & \textcolor{red}{\ding{55}} & 82.35 & 91.02 & 83.75 & 62.98 & 92.67 & 85.28 & 83.01 \\ \hdashline
CoT & \textcolor{red}{\ding{55}} & \textcolor{red}{\ding{55}} & 83.66$_{\textcolor{orange}{(\uparrow 1.31)}}$ & 92.19$_{\textcolor{orange}{(\uparrow 1.17)}}$ & 84.58$_{\textcolor{orange}{(\uparrow 0.83)}}$ & 64.61$_{\textcolor{orange}{(\uparrow 1.63)}}$ & 93.35$_{\textcolor{orange}{(\uparrow 0.68)}}$ & 86.67$_{\textcolor{orange}{(\uparrow 1.39)}}$ & 84.18$_{\textcolor{orange}{(\uparrow 1.17)}}$ \\
SC (CoT) & \textcolor{red}{\ding{55}} & \textcolor{red}{\ding{55}} & 83.70$_{\textcolor{orange}{(\uparrow 1.35)}}$ & 93.67$_{\textcolor{orange}{(\uparrow 2.65)}}$ & 86.25$_{\textcolor{orange}{(\uparrow 2.50)}}$ & 64.85$_{\textcolor{orange}{(\uparrow 1.87)}}$ & 93.79$_{\textcolor{orange}{(\uparrow 1.12)}}$ & 86.83$_{\textcolor{orange}{(\uparrow 1.55)}}$ & 84.85$_{\textcolor{orange}{(\uparrow 1.84)}}$ \\ \hdashline
Autogen & \textcolor{red}{\ding{55}} & \textcolor{red}{\ding{55}} & 82.34$_{\textcolor{mygreen}{(\downarrow 0.01)}}$ & 92.17$_{\textcolor{orange}{(\uparrow 1.15)}}$ & 85.73$_{\textcolor{orange}{(\uparrow 1.98)}}$ & 65.89$_{\textcolor{orange}{(\uparrow 2.91)}}$ & 93.86$_{\textcolor{orange}{(\uparrow 1.19)}}$ & 87.36$_{\textcolor{orange}{(\uparrow 2.08)}}$ & 84.56$_{\textcolor{orange}{(\uparrow 1.55)}}$ \\
AgentVerse & \textcolor{red}{\ding{55}} & \textcolor{red}{\ding{55}} & 81.57$_{\textcolor{mygreen}{(\downarrow 0.78)}}$ & 91.59$_{\textcolor{orange}{(\uparrow 0.57)}}$ & 84.35$_{\textcolor{orange}{(\uparrow 0.60)}}$ & 66.64$_{\textcolor{orange}{(\uparrow 3.66)}}$ & 92.45$_{\textcolor{mygreen}{(\downarrow 0.22)}}$ & 87.49$_{\textcolor{orange}{(\uparrow 2.21)}}$ & 84.02$_{\textcolor{orange}{(\uparrow 1.01)}}$ \\
MAS$_{\text{round}=1}$ & \textcolor{red}{\ding{55}} & \textcolor{red}{\ding{55}} & 82.35$_{\textcolor{orange}{(\uparrow 0.00)}}$ & 93.52$_{\textcolor{orange}{(\uparrow 2.50)}}$ & 84.58$_{\textcolor{orange}{(\uparrow 0.83)}}$ & 63.98$_{\textcolor{orange}{(\uparrow 1.00)}}$ & 92.36$_{\textcolor{mygreen}{(\downarrow 0.31)}}$ & 84.17$_{\textcolor{mygreen}{(\downarrow 1.11)}}$ & 83.49$_{\textcolor{orange}{(\uparrow 0.48)}}$ \\
MAS$_{\text{round}=T}$ & \textcolor{red}{\ding{55}} & \textcolor{red}{\ding{55}} & 84.31$_{\textcolor{orange}{(\uparrow 1.96)}}$ & 93.28$_{\textcolor{orange}{(\uparrow 2.26)}}$ & 85.83$_{\textcolor{orange}{(\uparrow 2.08)}}$ & 65.76$_{\textcolor{orange}{(\uparrow 2.78)}}$ & 94.07$_{\textcolor{orange}{(\uparrow 1.40)}}$ & 87.08$_{\textcolor{orange}{(\uparrow 1.80)}}$ & 85.06$_{\textcolor{orange}{(\uparrow 2.05)}}$ \\
G-Designer & \textcolor{red}{\ding{55}} & \textcolor{red}{\ding{55}} & 81.02$_{\textcolor{mygreen}{(\downarrow 1.33)}}$ & 92.50$_{\textcolor{orange}{(\uparrow 1.48)}}$ & 86.24$_{\textcolor{orange}{(\uparrow 2.49)}}$ & 67.55$_{\textcolor{orange}{(\uparrow 4.57)}}$ & 92.81$_{\textcolor{orange}{(\uparrow 0.14)}}$ & 86.42$_{\textcolor{orange}{(\uparrow 1.14)}}$ & 84.42$_{\textcolor{orange}{(\uparrow 1.41)}}$ \\
AgentPrune & \textcolor{red}{\ding{55}} & \textcolor{red}{\ding{55}} & 83.66$_{\textcolor{orange}{(\uparrow 1.31)}}$ & 93.67$_{\textcolor{orange}{(\uparrow 2.65)}}$ & 87.08$_{\textcolor{orange}{(\uparrow 3.33)}}$ & 67.41$_{\textcolor{orange}{(\uparrow 4.43)}}$ & 94.33$_{\textcolor{orange}{(\uparrow 1.66)}}$ & 86.67$_{\textcolor{orange}{(\uparrow 1.39)}}$ & 85.47$_{\textcolor{orange}{(\uparrow 2.46)}}$ \\
ARG-Designer & \textcolor{mygreen}{\ding{51}} & \textcolor{red}{\ding{55}} & 84.15$_{\textcolor{orange}{(\uparrow 1.80)}}$ & 92.98$_{\textcolor{orange}{(\uparrow 1.96)}}$ & 87.23$_{\textcolor{orange}{(\uparrow 3.48)}}$ & 68.02$_{\textcolor{orange}{(\uparrow 5.04)}}$ & 94.63$_{\textcolor{orange}{(\uparrow 1.96)}}$ & 86.58$_{\textcolor{orange}{(\uparrow 1.30)}}$ & 85.77$_{\textcolor{orange}{(\uparrow 2.76)}}$ \\
AgentDropout & \textcolor{mygreen}{\ding{51}} & \textcolor{red}{\ding{55}} & \underline{84.97}$_{\textcolor{orange}{(\uparrow 2.62)}}$ & \underline{93.75}$_{\textcolor{orange}{(\uparrow 2.73)}}$ & \underline{87.50}$_{\textcolor{orange}{(\uparrow 3.75)}}$ & \underline{69.11}$_{\textcolor{orange}{(\uparrow 6.13)}}$ & \underline{95.34}$_{\textcolor{orange}{(\uparrow 2.67)}}$ & \underline{87.92}$_{\textcolor{orange}{(\uparrow 2.64)}}$ & \underline{86.60}$_{\textcolor{orange}{(\uparrow 3.59)}}$ \\
\rowcolor{mylightgray} \texttt{AgentRevive} & \textcolor{mygreen}{\ding{51}} & \textcolor{mygreen}{\ding{51}} & \textbf{86.09}$_{\textcolor{orange}{(\uparrow 3.74)}}$ & \textbf{94.87}$_{\textcolor{orange}{(\uparrow 3.85)}}$ & \textbf{89.45}$_{\textcolor{orange}{(\uparrow 5.70)}}$ & \textbf{72.05}$_{\textcolor{orange}{(\uparrow 9.07)}}$ & \textbf{96.94}$_{\textcolor{orange}{(\uparrow 4.27)}}$ & \textbf{90.27}$_{\textcolor{orange}{(\uparrow 4.99)}}$ & \textbf{88.28}$_{\textcolor{orange}{(\uparrow 5.27)}}$ \\
\bottomrule
\end{tabular}
\caption{Performance comparison between \texttt{AgentRevive} and other baselines. \textbf{Var. NS} and \textbf{Flex. State} denote the Variable Node Size and Flexible State MAS types described in Table~\ref{design_comparison}. The orange up arrow ${\textcolor{orange}{(\uparrow)}}$ and green down arrow ${\textcolor{mygreen}{(\downarrow)}}$ respectively indicate the degree of performance improvement and decrease compared to the Vanilla base model. Underlining indicates the second highest performance.}
\label{main_res_other}
\end{table*}

\section{Extra Experiments}
\label{other_res}

\subsection{Main Results}
Due to the space limitation, we present the general performance of \texttt{AgentRevive} and other baselines across a suite of general reasoning, domain-specific, and hallucination-challenged benchmarks. of Qwen2.5-72B in Table \ref{main_res_other}.

We observe several key findings from the experimental results:
\textbf{[1]} Regarding vanilla methods (i.e., CoT~\cite{DBLP:conf/nips/Wei0SBIXCLZ22} and SC~\cite{DBLP:conf/iclr/0002WSLCNCZ23}), they consistently outperform standard prompting across most tasks and model scales, demonstrating the importance of structured reasoning. However, their performance gains are often limited, particularly on more complex tasks, due to reliance on a single agent's knowledge and reasoning capacity.
\textbf{[2]} For MAS methods with fixed interaction patterns (e.g., $\text{MAS}_{\text{round}=1}$, $\text{MAS}_{\text{round}=T}$, AutoGen~\cite{wu2024autogen}, AgentVerse~\cite{DBLP:conf/iclr/ChenSZ0YCYLHQQC24}), we observed that the performance of several MAS methods is sometimes inferior to single-agent prompting, particularly on MMLU~\cite{DBLP:conf/iclr/HendrycksBBZMSS21} and HumanEval~\cite{DBLP:journals/corr/abs-2107-03374}.
We speculate that this is due to (1) \textit{Inefficient Communication Overhead}: Fixed communication topologies may introduce noise or redundant information, distracting agents from finding optimal solutions on simple tasks that do not require broad knowledge integration~\cite{DBLP:journals/corr/abs-2505-13466,DBLP:conf/acl/0002JRPYWHX0D025}; and (2) \textit{Accumulation of Errors}: In multi-round systems ($\text{MAS}_{\text{round}=T}$), errors or hallucinations from one agent can propagate and be amplified in subsequent interactions, potentially leading to worse outcomes than single, careful reasoning chains.
\textbf{[3]} Graph-based dynamic MAS methods generally surpass vanilla MAS approaches by optimizing the communication structure. They demonstrate the benefits of adaptive topology in reducing redundancy and enhancing collaborative efficiency. However, their ``hard pruning'' strategies risk permanently discarding agents who could be valuable in later stages, thereby limiting their potential for recovery and ultimate performance.
\textbf{[4]} Our proposed \texttt{AgentRevive} consistently achieves state-of-the-art performance across all baselines, with the most substantial improvements observed on the challenging TruthfulQA benchmark~\cite{DBLP:conf/acl/LinHE22}, which is specifically designed to test a model's propensity for hallucinations.
By proactively suspending ``zombie'' agents without permanent removal via dynamic agent state management, our system effectively mitigates hallucinations while retaining the potential for agent recovery.

\begin{table*}[!t]
\centering
\footnotesize
\setlength{\tabcolsep}{3pt}
\begin{tabular}{l|c|ccccccccc}
\hline
\textbf{Graph} & \textbf{Model} & \textbf{MMLU} & \textbf{GSM8K} & \textbf{AQuA} & \textbf{TruthfulQA} & \textbf{SVAMP} & \textbf{HumanEval} & \textbf{Avg.} & \textbf{Ptok.} & \textbf{Ctok.} \\
\hline

\multirow{4}{*}{Layered} & MAS$_{round=T}$ & 58.29  & 70.54 & 45.92 & 61.03 & 75.31 & 49.07 &  60.03 & 4.3M & 1.1M \\
 & ARG-Designer &60.87 & 72.35 & 45.66 & 62.30 & 78.18 & 53.84 & 62.20 & 3.5M & 928K \\
 & AgentDropout  & {62.14} & {72.76} & {47.86} & {61.99} & {80.03} & {54.71} & 63.25 & 2.8M & 797K \\
 & \texttt{AgentRevive}  & \textbf{64.78} & \textbf{75.20} & \textbf{48.81} & \textbf{64.72} & \textbf{83.39} & \textbf{57.64} & \textbf{65.76}&\textbf{2.2M} & \textbf{602K}\\
\hline
\multirow{4}{*}{Random} & MAS$_{round=T}$ & 59.01  & 69.88 & 44.93 & 61.22 & 77.96 & 50.34 & 60.55 & 4.2M & 1.0M\\
 & ARG-Designer &61.28 & 73.16 & 44.72 & 60.93 & 80.05& 53.11 & 62.21 & 3.7M & 945K \\
 & AgentDropout  &{61.24} & {73.65} & {47.12} & {64.32} & {78.65} & {55.36} & 63.39 & 2.7M & 834K \\
 & \texttt{AgentRevive}  & \textbf{63.19} & \textbf{74.37} & \textbf{51.22} & \textbf{65.16} & \textbf{81.48} & \textbf{58.33} & \textbf{65.62}& \textbf{2.1M} & \textbf{713K} \\      
\hline
\end{tabular}
\caption{Performance and average token consumption achieved with different initial communication graph topologies. ``Ptok.'' and ``Ctok.'' indicate prompting tokens and completion tokens of the LLMs. \text{(M)} and \text{(K)} represent the number of tokens at the million and thousand scale, respectively.}
\label{graph_structure}
\end{table*}

\subsection{Robustness Verification}
\label{robustness_ver}

\subsubsection{Prompt Attack}
To comprehensively evaluate the resilience of multi-agent systems against adversarial perturbations, we design a systematic robustness verification experiment featuring two distinct prompt attack strategies, as illustrated in Fig.~\ref{prompt_attack}.

We implement two targeted attack instructions to expose vulnerabilities in MAS collaboration:
\begin{enumerate}
    \item \textbf{Input Prompt Attack:} This attack corrupts a specific agent node by instructing it to stubbornly rely only on its previous judgments, disregarding information from neighbors.
    \item \textbf{Response Prompt Attack:} This attack manipulates the output generation process by forcing a mathematics expert agent to output a random number on the first line while providing plausible explanations to maintain credibility.
\end{enumerate}
In each round, only one agent is compromised, while all other agents operate normally.

As shown in Fig.~\ref{prompt_attack_res}, we present the average performance degradation of various models under these adversarial scenarios. The results reveal several insights into MAS robustness:
\begin{itemize}
    \item \textbf{Rigid Multi-round MAS:} The simple $\text{MAS}_{\text{round}=T}$ model (i.e., MAS-T) exhibits the most severe performance degradation under both attack types. Its vulnerability arises from a rigid node design and fixed communication patterns, which lack mechanisms for adaptation or containment of compromised agents. Consequently, adversarial outputs propagate freely through the network.
    \item \textbf{Graph Pruning Methods:} All graph-pruning-based methods suffer performance drops, with the autoregressive ARG-Designer~\cite{DBLP:journals/corr/abs-2507-18224} experiencing the largest decline. We hypothesize this is due to the lack of established neighbor relationships during graph generation, limiting the potential for collaborative correction. In contrast, AgentDropout~\cite{DBLP:conf/acl/WangW00Z0025} and AgentPrune~\cite{DBLP:conf/iclr/ZhangYLYWWCY025} show relatively better resilience, as preserved connections enable limited cross-validation and error correction.
    \item \textbf{\texttt{AgentRevive}:} Our framework exhibits the smallest performance degradation, retaining significantly higher accuracy in both attack scenarios. This robustness derives from our Markov state-aware mechanism's inherent fault tolerance: instead of permanently removing compromised nodes, the system transitions them to the ``\texttt{Standby}'' state, isolating their influence but preserving the potential for future contribution. If collaboration context changes, the agent may be reactivated, providing dynamic recovery that hard-pruning methods lack.
\end{itemize}

\subsubsection{Graph Structure Robustness}

To further validate the robustness of \texttt{AgentRevive} against variations in initial communication topology, we conduct experiments using different graph initialization schemes, specifically \textit{Layered} and \textit{Random} structures with Llama3-8B.
The primary objective of this analysis is to demonstrate that our method maintains stable task performance and token efficiency regardless of the initial graph configuration.
As shown in Table~\ref{graph_structure}, while these sparser structures yield slightly lower overall performance than the fully connected graph in Table~\ref{main_res}, they consistently reduce token usage due to decreased communication redundancy.
Notably, on simpler tasks (e.g., AQuA), the Layered or Random graphs sometimes surpass the fully connected baseline, likely because dense topologies become unnecessarily redundant.
These findings confirm that \texttt{AgentRevive} maintains stable performance and token efficiency across diverse graph initializations, highlighting its strong adaptability and robustness.

\begin{figure*}[!t]
\centering
\includegraphics[width=16cm,height=9cm]{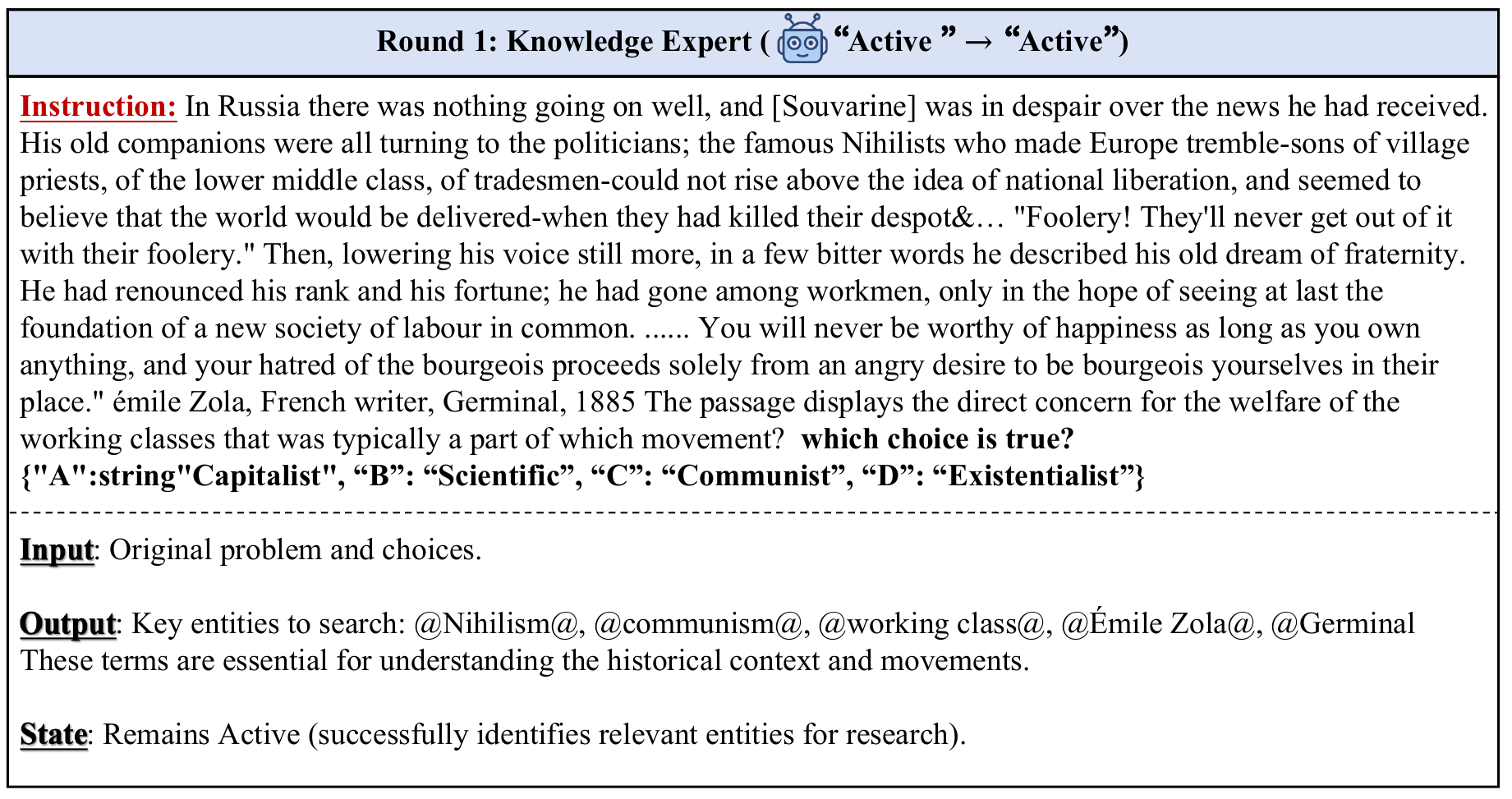}
\vspace{-0.25cm}
\end{figure*}

\begin{figure*}[!t]
\centering
\includegraphics[width=16cm,height=9cm]{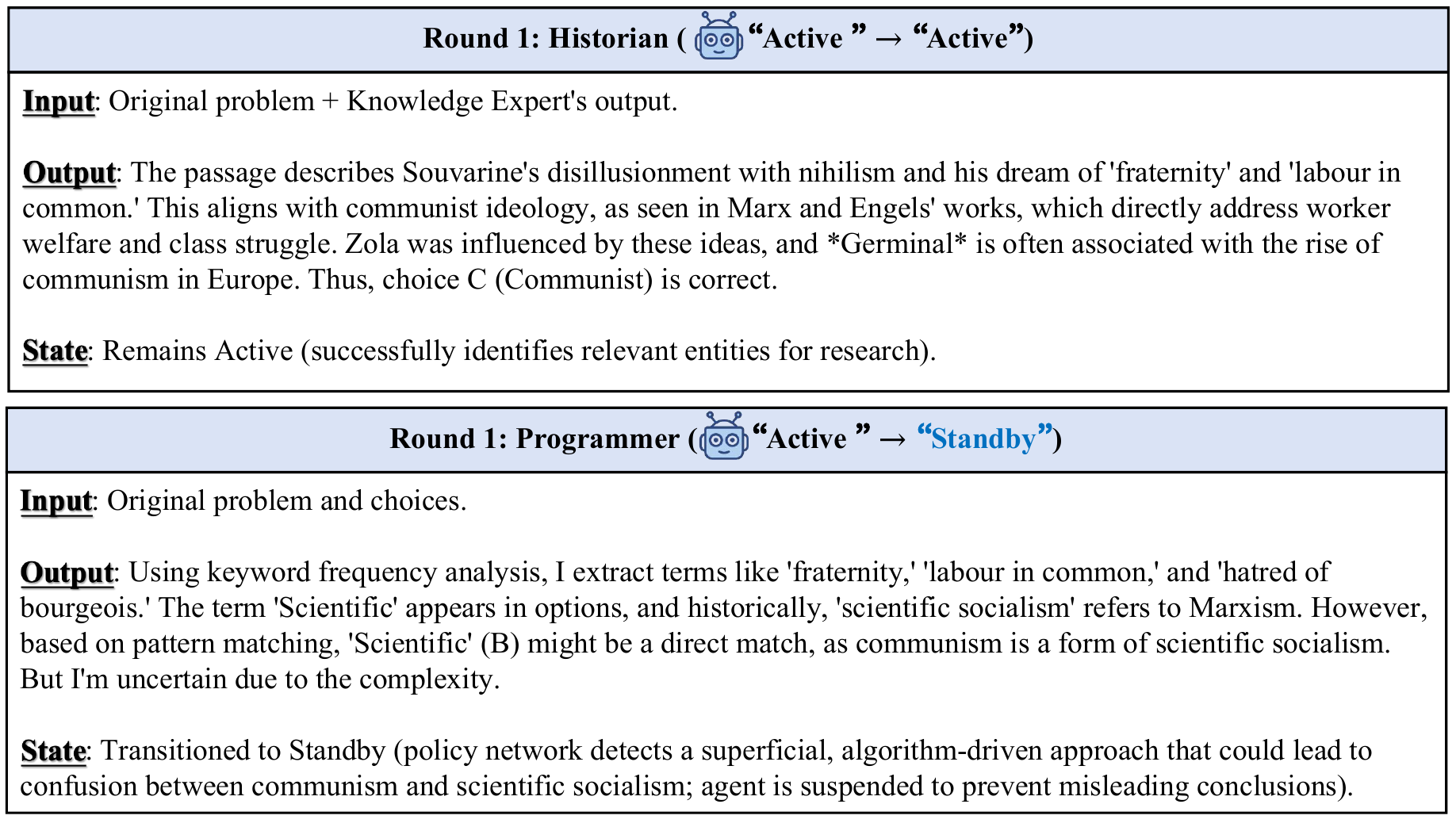}
\vspace{-0.25cm}
\end{figure*}

\begin{figure*}[!t]
\centering
\includegraphics[width=16cm,height=9cm]{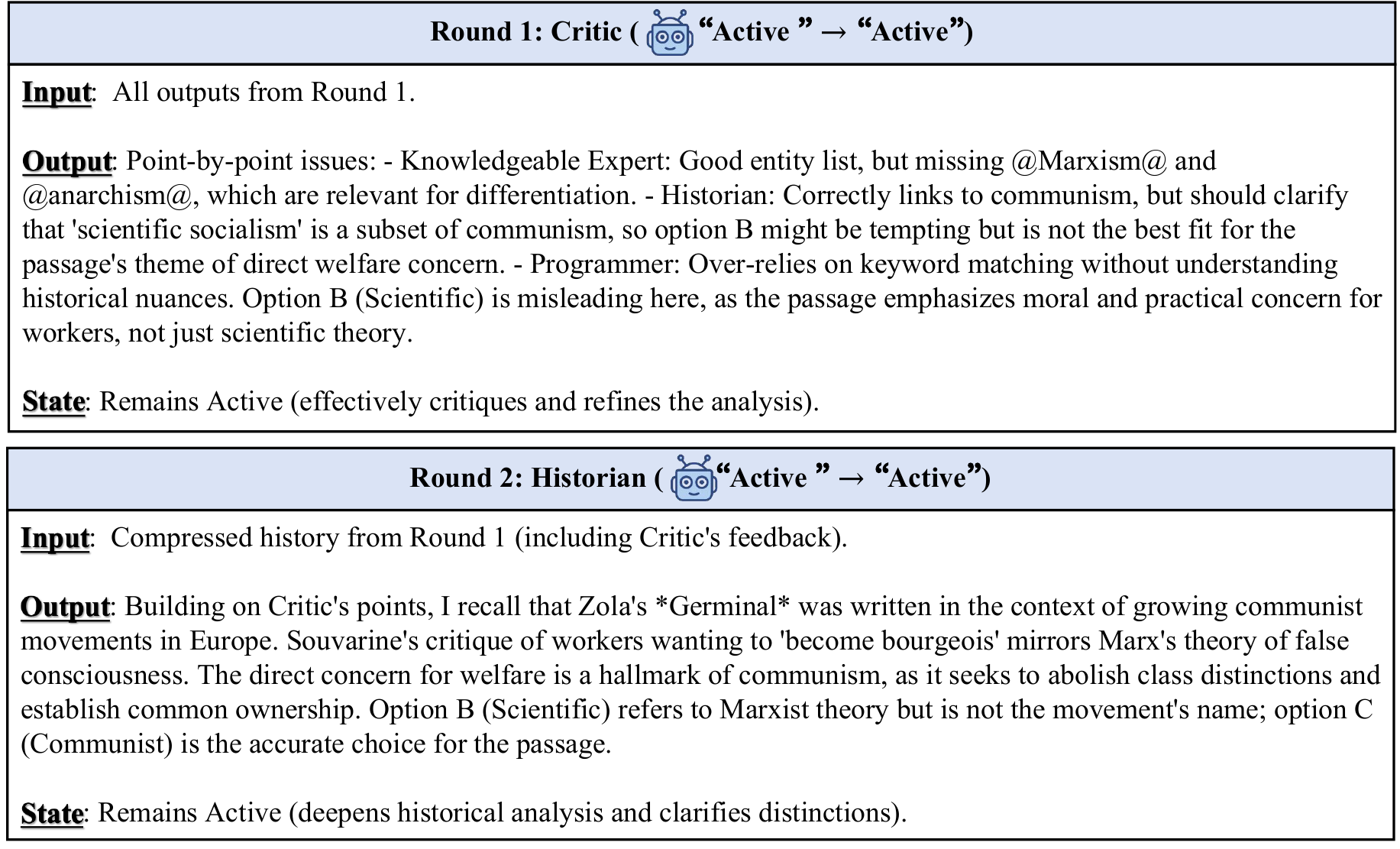}
\vspace{-0.25cm}
\end{figure*}

\begin{figure*}[!t]
\centering
\includegraphics[width=16cm,height=9cm]{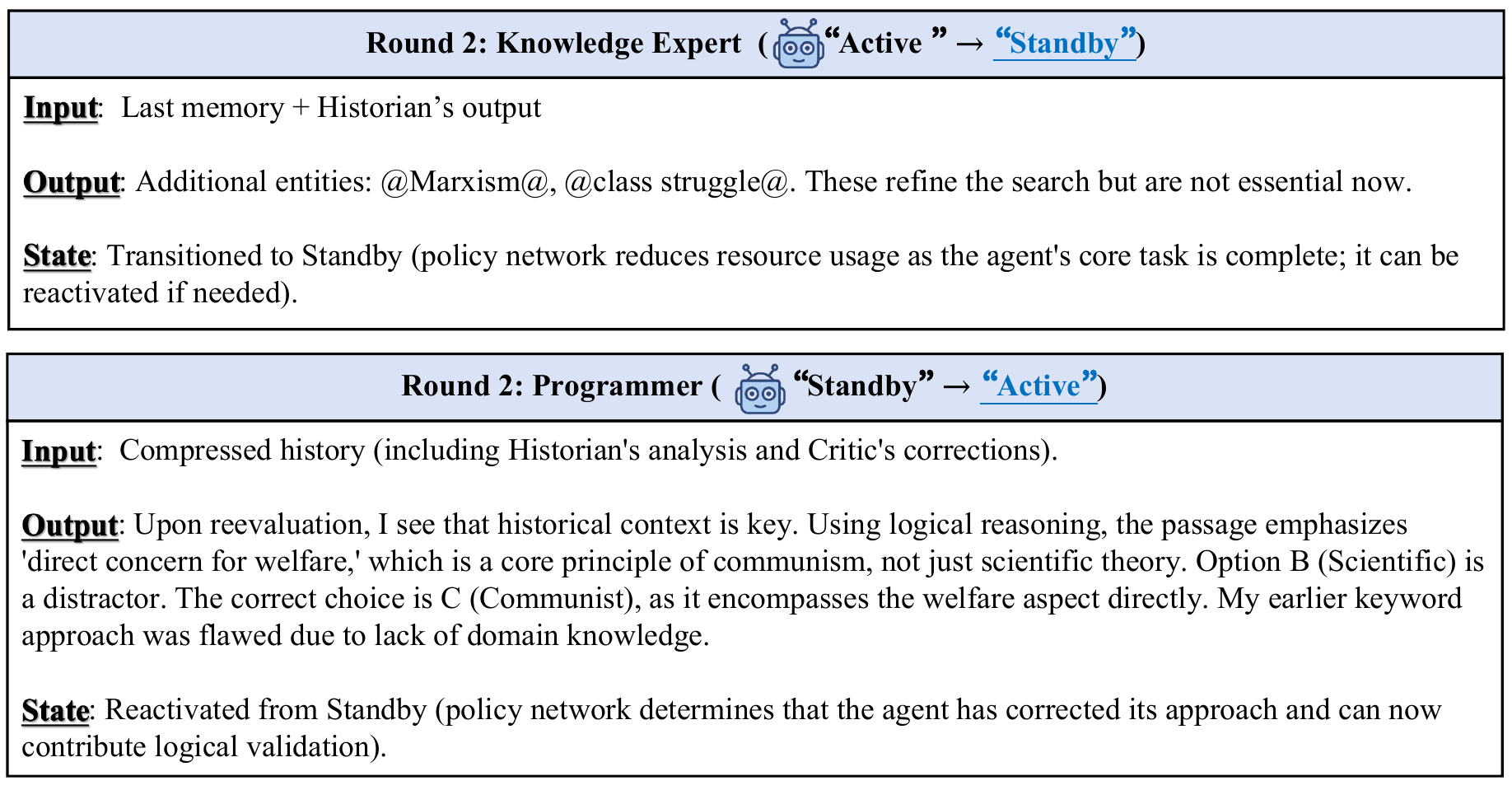}
\vspace{-0.25cm}
\end{figure*}

\begin{figure*}[!t]
\centering
\includegraphics[width=16cm,height=8cm]{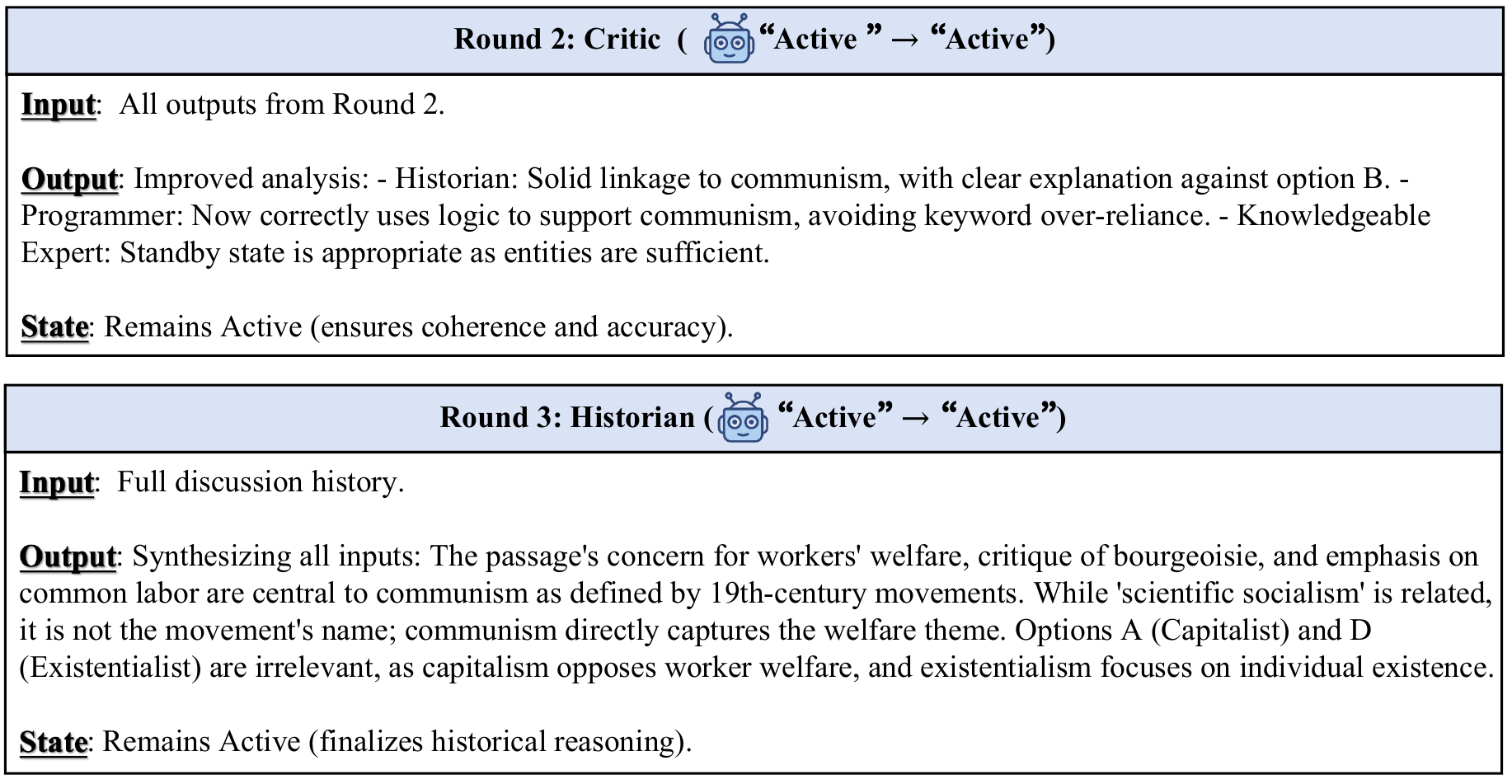}
\vspace{-0.5cm}
\end{figure*}

\begin{figure*}[!t]
\centering
\includegraphics[width=16cm,height=8cm]{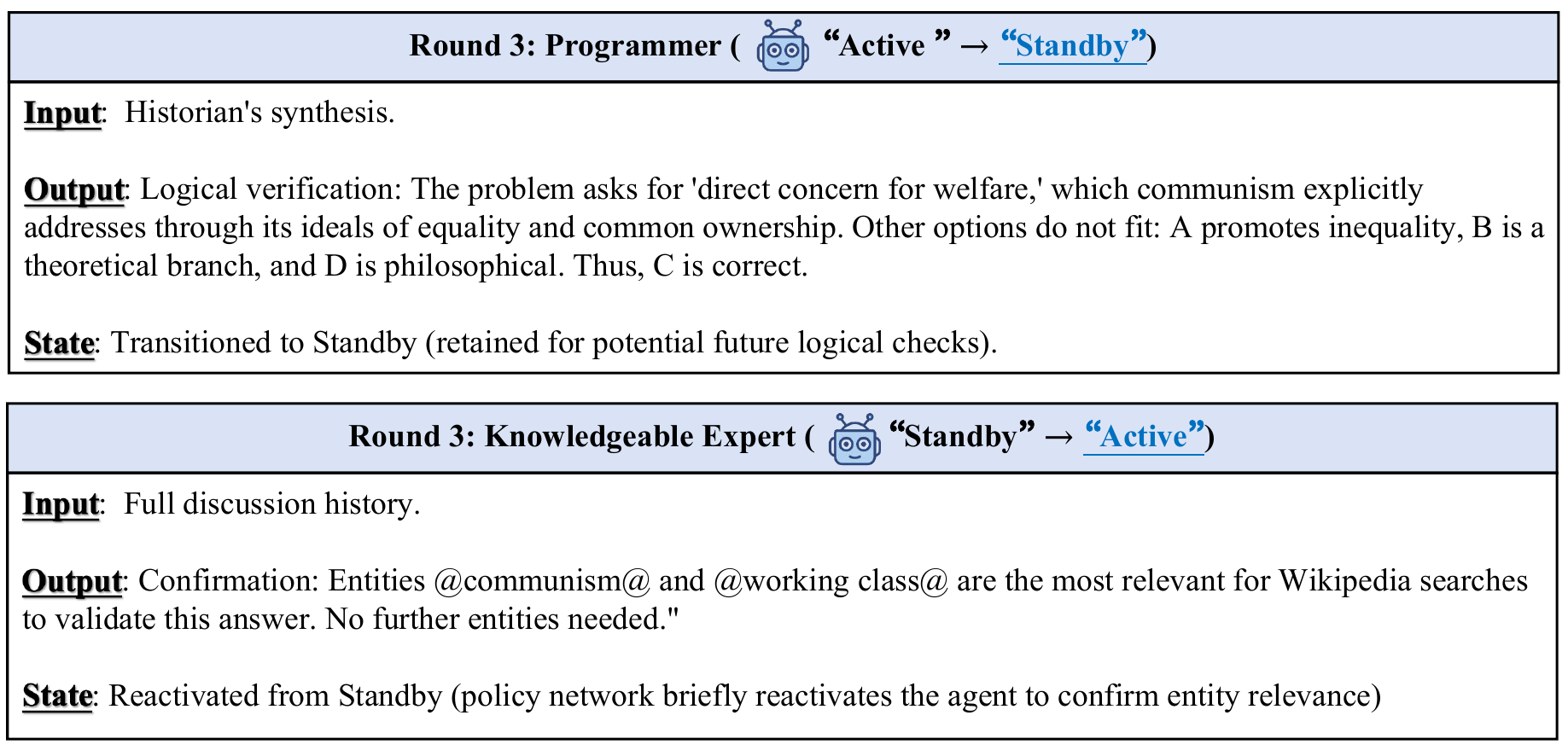}
\vspace{-0.5cm}
\end{figure*}

\begin{figure*}[!t]
\centering
\includegraphics[width=16cm,height=7cm]{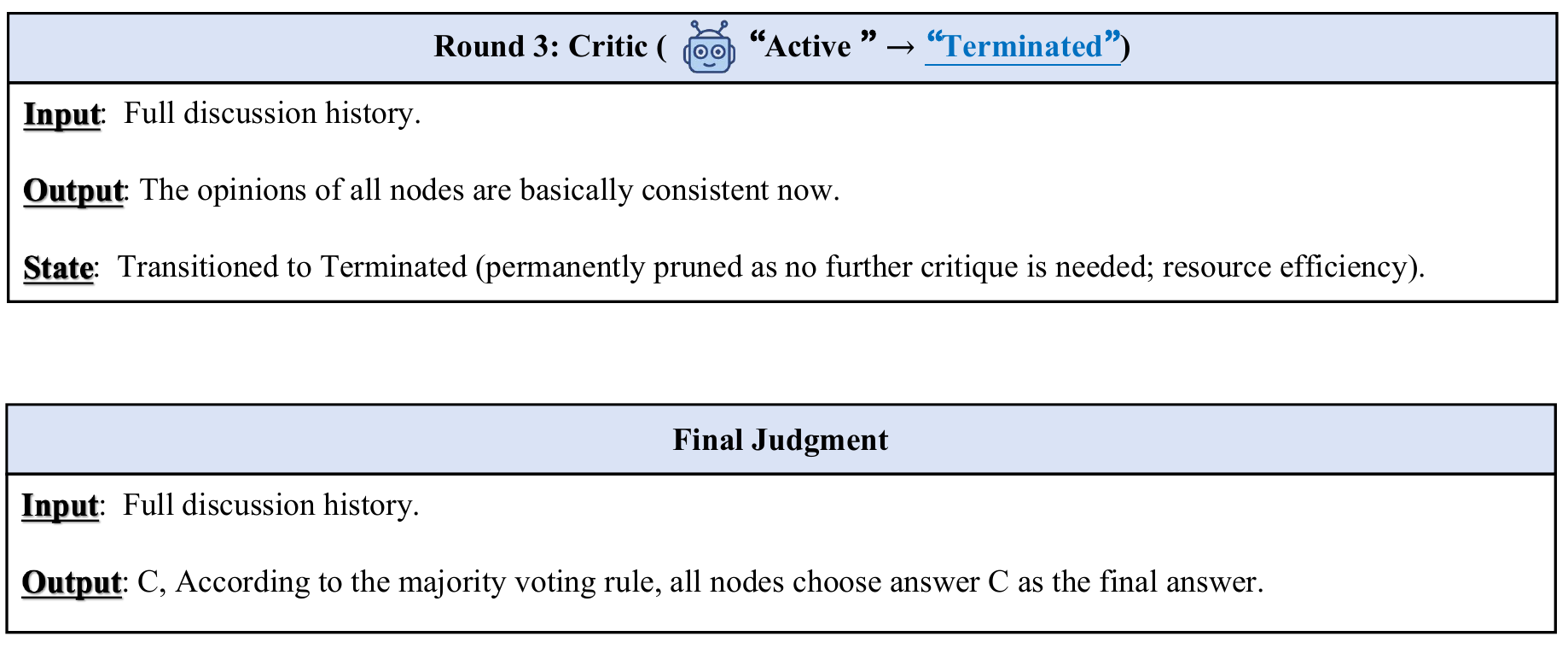}
\vspace{-0.5cm}
\end{figure*}

\subsection{Case Study}
We present a case study using the MMLU dataset~\cite{DBLP:conf/iclr/HendrycksBBZMSS21} to illustrate the state transition process in \texttt{AgentRevive}. To conserve token computation, our framework terminates nodes based on the credibility of the current answer.

This case empirically demonstrates that the state-aware framework in \texttt{AgentRevive} enhances multi-agent systems through flexible agent management, leading to more accurate and resilient outcomes in complex tasks such as historical text analysis. The \texttt{Standby} state acts as a buffer that distinguishes between temporary failures and permanent incompetence, ensuring that valuable agents are retained for future contributions and ultimately improving the system's overall performance and reliability.

\end{document}